\documentclass[11pt]{article}
\usepackage[final]{acl}

\usepackage{times}
\usepackage{latexsym}

\usepackage[T1]{fontenc}

\usepackage[utf8]{inputenc}
\usepackage{amsfonts}

\usepackage{microtype}

\usepackage{inconsolata}

\usepackage{graphicx}

\usepackage{pifont}
\usepackage{makecell}
\usepackage{multirow}
\usepackage{amsmath}
\usepackage{subcaption}
\usepackage{booktabs}
\usepackage{xspace}
\usepackage{placeins}
\usepackage{dsfont}
\usepackage{graphicx}

\usepackage{xcolor}
\usepackage{xspace}
\definecolor{purple(munsell)}{rgb}{0.62, 0.0, 0.77}

\newcolumntype{L}[1]{>{\raggedright\let\newline\\\arraybackslash\hspace{0pt}}p{#1}}
\newcolumntype{M}[1]{>{\raggedright\let\newline\\\arraybackslash\hspace{0pt}}m{#1}}
\newcolumntype{C}[1]{>{\centering\arraybackslash\hspace{0pt}}p{#1}}

\usepackage{listings}
\usepackage[most]{tcolorbox}
\tcbuselibrary{listings}

\lstdefinestyle{cascadepromptstyle}{
  basicstyle=\ttfamily\footnotesize,
  breaklines=true,
  breakatwhitespace=false,
  columns=fullflexible,
  keepspaces=true,
  showstringspaces=false,
  upquote=true,
  literate={\ }{{\ }}1,
}

\newtcblisting{promptbox}[1]{
  enhanced, breakable, listing only,
  colback=blue!3, colframe=blue!50!black,
  fonttitle=\bfseries\small, title=#1,
  boxrule=0.4pt, arc=2pt, left=4pt, right=4pt, top=2pt, bottom=2pt,
  listing options={style=cascadepromptstyle},
}

\newtcblisting{codebox}[1]{
  enhanced, breakable, listing only,
  colback=gray!4, colframe=gray!55!black,
  fonttitle=\bfseries\small, title=#1,
  boxrule=0.4pt, arc=2pt, left=4pt, right=4pt, top=2pt, bottom=2pt,
  listing options={style=cascadepromptstyle},
}

\newcommand{\our}{\textsc{EDGE}\xspace}

\definecolor{deltaup}{HTML}{1B7F3A}
\definecolor{deltadn}{HTML}{B91C1C}
\newcommand{\dup}[1]{\textcolor{deltaup}{#1}}
\newcommand{\ddn}[1]{\textcolor{deltadn}{#1}}

\title{\our: Error Dependency Graph-Guided Multi-Error Attribution in Multi-Agent LLM Systems}

\author{Jun Hou \quad Priya Pitre \quad Yi Fang \quad Xuan Wang \\
  Virginia Tech \\
  \texttt{\{junh, priyapitre, yif, xuanw\}@vt.edu}}

\begin{document}
\maketitle

\begin{abstract}
Large language model (LLM) agent failures often contain multiple related errors rather than a single mistake. Existing attribution methods usually identify a responsible agent, step, or root cause, but do not explicitly model dependency between errors. We introduce \our~\footnote{Code and experimental artifacts are available at \url{https://github.com/JuneHou/EDGE}.}, an \textbf{E}rror \textbf{D}ependency \textbf{G}raph-guided multi-\textbf{E}rror attribution framework. \our constructs an error dependency graph from observed error events and validates a reliable causal subset through counterfactual rollout. The inference graph guides a two-stage LLM-as-judge detector for error attribution, and the intervention-validated subgraph provides a more reliable basis for explanation and repair analysis. Experiments on TRAIL and MAST show that \our improves category-level multi-error attribution across most evaluated models and settings. Experiments with adapted Who\&When-style prompts show that the graph helps across prompting strategies. These results suggest that dependency structure is a useful diagnostic prior for agent failures beyond isolated root-cause prediction.
\end{abstract}

\section{Introduction}

Large language model (LLM) agents are increasingly used for complex multi-step tasks that involve planning, retrieval, tool use, and reasoning. For failed executions, the final outcome indicates task failure, but rarely explains how the failure developed. Recent work has therefore moved toward trace-level failure attribution, which identifies the step and error behind a failed execution~\cite{zhang2025which,barke2026agentrx,ma2025automatic,wang2026flat,zhu2026raffles}. These methods center on single-error attribution and do not provide full step-level multi-error annotation.

\begin{figure}[t]
  \centering
  \includegraphics[width=\linewidth]{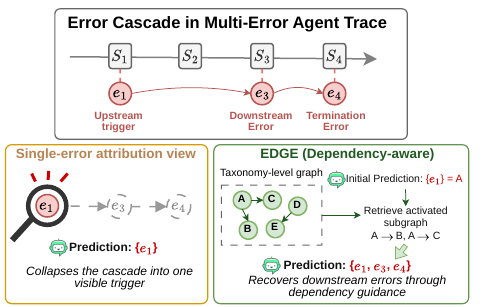}
  \caption{Errors in agent traces form dependency cascades
    ($e_1\!\to\!e_3\!\to\!e_4$). Unlike single-locus attribution,
    \our detects the downstream errors via a taxonomy-level
    dependency graph.}
  \label{fig:motivation}
\end{figure}

TRAIL reflects a newer direction by providing fine-grained span-level annotations for multiple errors in each failed trace~\cite{deshpande2025trail}. Within a failed trace, errors play distinct structural roles, with some as upstream triggers, others as downstream effects, and others as parallel symptoms (Figure~\ref{fig:motivation}). These relations recur across traces, motivating a dependency graph constructed from the full corpus. When extended for multi-label output, single-error methods are inefficient and treat each error as an isolated target without modeling how one error makes another more likely. We therefore study multi-error attribution as dependency-aware diagnosis, where the dependency structure guides efficient and effective attribution and the validated subset supports explanation.

To achieve this goal, we frame multi-error attribution as dependency-aware diagnosis. We use within-trace ordering and cross-trace regularity to capture how error categories recur and propagate across failed executions. Based on this view, we introduce \textbf{\our}, an error dependency graph-guided framework for multi-error attribution in LLM agent traces. \our constructs an error dependency graph from observed error events, validates selected edges through counterfactual rollout, and uses the resulting graph to guide a two-stage detector. The first stage predicts an initial error set, and the second stage verifies downstream errors suggested by the graph. Score-filtered observational edges improve detection coverage, while intervention-validated edges provide a more reliable structure for explanation and repair analysis. This role separation lets \our use observational edges to broaden attribution coverage when intervention evidence is limited, while preserving the causal interpretability of the validated subset.

\begin{figure*}[!t]
  \centering
  \includegraphics[width=\textwidth]{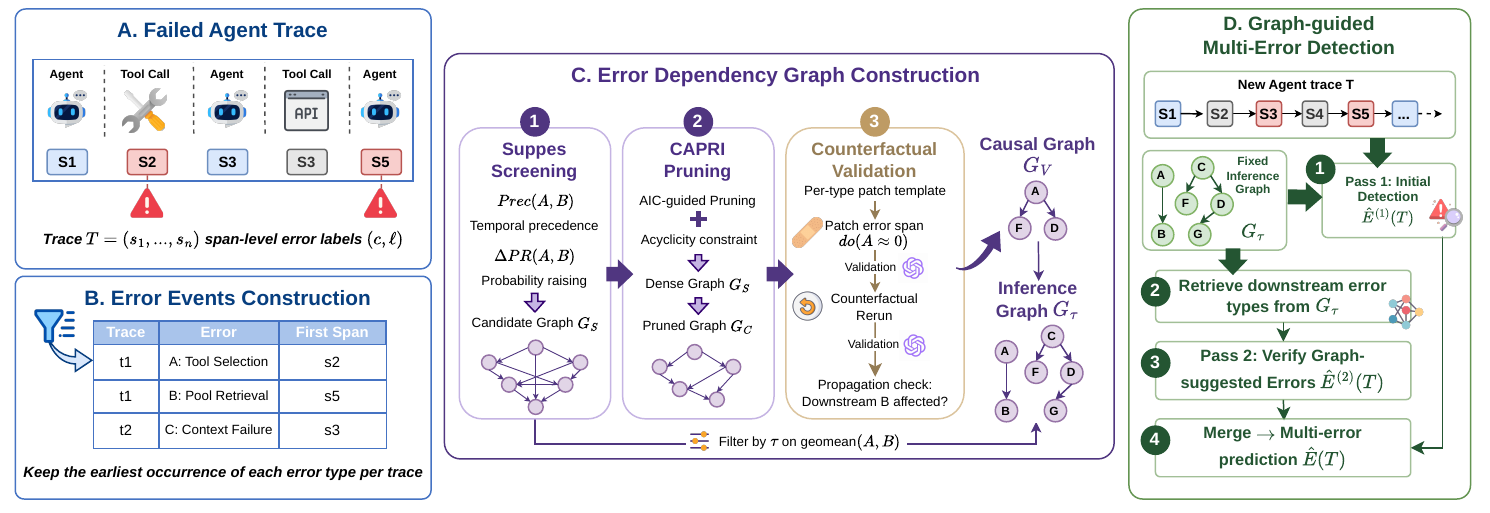}
  \caption{Overview of \our.
    \textbf{(Left)} An observational correlation graph $\mathcal{G}_S$ and a
    sparse causal-candidate graph $\mathcal{G}_C$ are inferred from event
    statistics via Suppes screening and CAPRI pruning.
    \textbf{(Middle)} Causal-candidate edges in $\mathcal{G}_C$ are validated
    by counterfactual rollout, yielding $\mathcal{G}_V$; score-filtered
    correlation edges from $\mathcal{G}_S$ are unioned with $\mathcal{G}_V$
    to form $\mathcal{G}_\tau$.
    \textbf{(Right)} Two-stage detection: Stage~1 predicts with
    full-graph context; Stage~2 verifies downstream categories
    reachable from Stage-1 predictions.}
  \label{fig:pipeline}
\end{figure*}

We evaluate \our on TRAIL~\cite{deshpande2025trail} and MAST~\cite{cemri2026multi}, two benchmarks with distinct error taxonomies that cover span-level and trace-level multi-error attribution settings, demonstrating that the pipeline is taxonomy-agnostic. Across these benchmarks, \our improves category-level attribution for most evaluated models and settings. Additional experiments with adapted Who\&When-style prompts show that the inferred graph can guide different attribution formats, rather than only one detector prompt. Our contributions are threefold. \underline{First}, we formulate multi-error attribution as dependency-aware diagnosis for LLM agent traces. \underline{Second}, we construct an error dependency graph from observed error events and inject it into a two-stage LLM-as-judge detector. \underline{Third}, we separate prediction from explanation using $\mathcal{G}_\tau$ for attribution coverage and the intervention-validated subgraph $\mathcal{G}_V$ for repair and propagation analysis.

\section{Related Work}

\paragraph{Failure Attribution in Agent Systems}
Error attribution in agent systems studies how to identify the step, agent, or error instance most responsible for a failed execution trace. This work focuses on the LLM-reasoning paradigm emphasized in recent surveys of agent trajectory analysis~\cite{wang2026survey}.
Early reasoning-based attribution methods explore how the LLM reads and organizes the trace. Who\&When~\cite{zhang2025which} formulates attribution as direct trajectory reasoning, while other methods improve this paradigm through hierarchical context, iterative adjudication, and structured causal prompting~\cite{banerjee2025did,zhu2026raffles,west2509abduct}. AgentErrorBench and AgentDebug further introduce a modular agent-error taxonomy, annotated failure trajectories, and attribution-oriented debugging from root-cause feedback~\cite{zhu2509llm}.
Recent work has further introduced causal structure into LLM-based attribution. CDC-MAS~\cite{ma2025automatic} uses multi-granularity causal inference, CHIEF~\cite{wang2026flat} prompts an LLM to generate hierarchical causal graphs with counterfactual screening, and AgentRx~\cite{barke2026agentrx} combines constraint-based evidence with LLM judgment for more auditable diagnosis. These attribution and debugging methods move beyond surface-level reasoning, but they still primarily target a single decisive error, critical step, or local root cause behind the final failure outcome. \our instead models failed traces as multi-error propagation, where earlier errors can make later errors more likely through an inferred dependency structure over error categories.

\paragraph{Sparse Event Progression Modeling}
Learning propagation structure from event data studies how earlier events are statistically or causally associated with later events in accumulative or cascading processes. Some work learns temporal influence and causal structure from timestamped events using Hawkes processes, temporal point processes, or Granger-style predictability~\cite{Cai2021THPsTHA,Qiao2023StructuralHPB,Wu2024LearningGCV}, and recent representation-based methods further learn temporal causal structures through latent regimes, sample-level relations, multimodal representations, or causal sequence prediction~\cite{Chen2025TowardTCD,Rahmani2023CastorCTK,Math2026ScalableSCE,Li2025InterpretableCWG}. However, agent-error traces are sparse, partially ordered, and semantic rather than dense continuous streams, so intensity-based event progression learning is suboptimal here. Representation-based approaches additionally require data and modeling assumptions that do not fit compact LLM-detector guidance~\cite{Constantinou2021TheIOC}.
Suppes-style accumulative progression modeling was applied to cancer progression by CAPRESE~\cite{loohuis2014inferring} and CAPRI~\cite{ramazzotti2015capri}, with TRONCO providing a modular implementation~\cite{desano2016tronco}. It constructs correlation-based dependency hypotheses from sparse events using temporal priority and probability raising. \our adopts this view to infer category-level propagation structure from recurring error-event patterns across traces.

\section{Method}

\our models error propagation in multi-error agent traces. Given span-level annotations of error event types and locations, we first formalize the multi-error attribution task, then construct an observational correlation graph $\mathcal{G}_S$ and a causal-candidate graph $\mathcal{G}_C$ over error categories, validate a trustworthy causal anchor with targeted interventions, and finally build an inference graph that augments this anchor with score-filtered correlation edges (Figure~\ref{fig:pipeline}). Section~\ref{problem} formalizes the task and notation. Section~\ref{candidate_graph} describes the two observational graphs ($\mathcal{G}_S$, $\mathcal{G}_C$) built from temporal precedence and probability raising. Section~\ref{graph_validation} introduces intervention-based edge validation. Section~\ref{graph_guided_detection} presents how the selected propagation graph is incorporated into graph-guided error detection. We use $\mathcal{G}_\tau$ (validated $\cup$ score-filtered correlation from $\mathcal{G}_S$) as the default detector input and additionally report $\mathcal{G}_V$ (validated only) as an ablation alternative. A complete symbol summary is provided in Appendix~\ref{app:notation}.

\subsection{Problem Setup} \label{problem}

Let a trace $T = (s_1, s_2, \ldots, s_N)$ denote a sequence of spans ordered by execution time. Each trace is annotated with a set of error events $E(T) = \{(c_j, \ell_j)\}$, where $c_j \in \mathcal{C}$ is an error category from an arbitrary multi-error taxonomy $\mathcal{C}$ and $\ell_j$ is the string identifier of the span at which that error first emerges, evaluated by exact match. Its position in the execution order $(s_1, \ldots, s_N)$ supplies the temporal ordering used internally by the candidate-graph construction (Section~\ref{candidate_graph}). The goal is to predict, for each trace, a set of category-location pairs $\hat{E}(T)$.

We represent error-category dependencies with a directed graph $\mathcal{G}=(\mathcal{C},\mathcal{R})$, where $\mathcal{R}$ is a set of directed propagation edges. Each node is an error category and each edge \(A\rightarrow B\) denotes a hypothesized propagation relation from source error \(A\) to downstream error \(B\). We distinguish observational propagation edges from intervention-validated edges, which are interpreted as empirically supported causal propagation links.

\subsection{Observational Graph Construction} \label{candidate_graph}

We adopt the Suppes-Bayes Causal Network construction of CAPRI~\cite{ramazzotti2015capri} (as implemented in the TRONCO package~\cite{desano2016tronco}), which first prima-facie-filters edges by Suppes' temporal-priority and probability-raising conditions~\cite{suppes1970probabilistic} and then prunes the Suppes-screened graph by a regularized likelihood score, to construct an observational correlation graph $\mathcal{G}_S$ and a causal-candidate graph $\mathcal{G}_C$ over error categories. The key adaptation is to treat error-category occurrences as progression events while using step annotations to derive category-level event variables for temporal-order estimation. Let $\mathcal{T}$ denote the set of annotated traces, then for each trace $T\in\mathcal{T}$ and category $A\in\mathcal{C}$, we define
\begin{equation}
\label{eq:x_a}
X_A(T)
=
\mathbf{1}\!\left[
\exists(c,\ell)\in E(T): c=A
\right],
\end{equation}
where $X_A(T)=1$ indicates that error category $A$ occurs anywhere in trace $T$, and we write $t_A(T)$ for the first-event index of $A$ (formal definition in Appendix~\ref{app:graph_construction}). Tied events are retained, but do not support strict temporal direction.

\paragraph{Suppes-based Candidate Screening.}
We first identify prima facie candidate edges using Suppes' probabilistic theory of causation~\cite{suppes1970probabilistic}.
For a directed pair \(A \rightarrow B\), we require both temporal priority and probability raising.
Let \(\mathcal{T}_{AB}\) denote the set of traces in which both \(A\) and \(B\) are present without ties.
Because our annotations provide event order within each trace, we estimate temporal priority by aggregating precedence evidence across co-occurring traces, and operationalize Suppes' probability-raising condition as the difference in $B$'s conditional occurrence rate:
\begin{equation}
\label{eq:prec}
\resizebox{0.85\linewidth}{!}{$
\mathrm{Prec}(A,B)
=
\frac{
\left|
\left\{
T\in \mathcal{T}_{AB} : t_A(T)<t_B(T)
\right\}
\right|
}{
|\mathcal{T}_{AB}|
}.
$}
\end{equation}

\begin{equation}
\label{eq:delta_pr}
\resizebox{0.85\linewidth}{!}{$
\begin{aligned}
\Delta_{\mathrm{PR}}(A,B) ={} & P(X_B{=}1 \mid X_A{=}1) \\
 & {}- P(X_B{=}1 \mid X_A{=}0).
\end{aligned}
$}
\end{equation}

We retain \(A \rightarrow B\) when both criteria exceed predefined thresholds with sufficient joint support (Appendix~\ref{app:graph_construction}).
The output of this stage is the \emph{observational correlation graph} \(\mathcal{G}_S=(\mathcal{C},\mathcal{R}_S)\), whose edges represent observational propagation hypotheses. For each surviving edge $A \rightarrow B \in \mathcal{R}_S$ (both criteria positive), we derive a propagation weight given by the geometric mean of the two screening scores,
\begin{equation}
\label{eq:w_s}
w_S(A,B) = \sqrt{\mathrm{Prec}(A,B)\cdot\Delta_{\mathrm{PR}}(A,B)},
\end{equation}
which is used for inference-time edge ranking (Section~\ref{graph_guided_detection}). Threshold is reported in Appendix~\ref{app:graph_construction}.

\paragraph{CAPRI-based Graph Pruning.}
The Suppes screen produces $\mathcal{G}_S$, but it may retain indirect or transitive dependencies. We therefore apply a CAPRI-style score-based pruning procedure to obtain a sparser \emph{causal-candidate graph}, selected from the search space of directed acyclic graphs (DAGs)
\begin{equation}
\label{eq:admissible}
\Omega(\mathcal{R}_S)
=
\left\{
\mathcal{G}:
\mathcal{G}\ \text{is a DAG and}\
\mathcal{R}\subseteq \mathcal{R}_S
\right\}.
\end{equation}
Each category node is modeled as a binary event conditioned on its parent set, using the occurrence variables \(X_A(T)\). We select the causal-candidate graph $\mathcal{G}_C=(\mathcal{C},\mathcal{R}_C)$ as the minimizer over $\Omega(\mathcal{R}_S)$ of an AIC-style score that combines the maximized log-likelihood of the binary category-event model with a parameter-count penalty. The full objective, optimization, and Suppes threshold values are deferred to Appendix~\ref{app:graph_construction}.

\subsection{Intervention-based Graph Validation}  \label{graph_validation}

To strengthen causal interpretation, we validate $\mathcal{G}_C$ edges with counterfactual interventions.

\paragraph{Controlled Counterfactual Rollout.}
For each candidate edge $A \rightarrow B \in \mathcal{R}_C$, we collect the trace set $\mathcal{T}_{A\prec B}$ where $t_A(T) < t_B(T)$. We intervene on the source span $s_{t_A(T)}$ using an LLM-based repair module that applies a category-specific patch template to rewrite the span's input or output and eliminate $A$. We then run the counterfactual rollout with all tool outputs held to their original values, so only the patched span and its downstream reasoning change. A repair verifier (Judge A) labels each rollout with $\mathrm{resolved}_A(T) \in \{0,1\}$ ($=1$ iff $A$ is removed). Only rollouts with $\mathrm{resolved}_A(T)=1$ enter effect estimation. Because tool outputs are held fixed, any downstream change in $B$ is attributable to the patch on $A$, so the effect estimate captures the \emph{controlled direct effect} of $A$ on $B$ along the agent's reasoning path~\cite{pearl2001direct}. Patch templates, postchecks, Judge-A prompts, and hyperparameters are in Appendix~\ref{app:graph_construction}.

\paragraph{Edge Effect Estimation and Validation.}
For each verified rollout, an effect evaluator (Judge B) labels whether $B$ remains present in the continuation. Let $X_B^{\mathrm{cf}}(T;A) \in \{0,1\}$ denote this Judge-B label. We estimate the downstream reduction effect of edge $A\rightarrow B$ as the boolean risk-difference
\begin{equation}
\label{eq:delta}
\resizebox{0.85\linewidth}{!}{$
\Delta(A \to B) = \mathbb{E}_{T\in\mathcal{T}^{\mathrm{ver}}_{A\prec B}}\!\bigl[X_B(T) - X_B^{\mathrm{cf}}(T;A)\bigr],
$}
\end{equation}
where $\mathcal{T}^{\mathrm{ver}}_{A\prec B} \subseteq \mathcal{T}_{A\prec B}$ retains only traces with $\mathrm{resolved}_A(T)=1$. We retain $A \rightarrow B$ to form $\mathcal{G}_V=(\mathcal{C},\mathcal{R}_V)$ when $\Delta(A \to B) > \tau_{\mathrm{val}}$ (Appendix~\ref{app:graph_construction}). The validated graph $\mathcal{G}_V$ serves two roles: one is the trustworthy causal anchor retained in all thresholded inference graphs, so observational edges augment rather than replace intervention evidence, and another is an interpretable propagation map for post-hoc explanation, repair prioritization, and system-level debugging.

\subsection{Graph-Guided Error Detection} \label{graph_guided_detection}

With both observational and intervention-validated propagation graphs, our downstream detector follows an LLM-as-judge format augmented with structured graph guidance. Given a trace $T$ and taxonomy $\mathcal{C}$, a baseline judge $\mathcal{J}$ receives the serialized trace together with the definitions of categories and predicts a set of errors.

\paragraph{Inference Graph Selection.}
The causal-only graph \(\mathcal{G}_V\) is maximally interpretable, but it may lack strong correlations that are visible observationally yet difficult to validate by intervention in available sample sizes. We therefore construct a thresholded inference graph by taking the union of score-filtered correlation edges from $\mathcal{G}_S$ and all intervention-validated causal edges:
\begin{equation}
\label{eq:r_tau}
\resizebox{0.85\linewidth}{!}{$
\mathcal{R}_{\tau}
=
\left\{
A\rightarrow B \in \mathcal{R}_S :
w_S(A,B)\geq \tau
\right\}
\cup
\mathcal{R}_V,
$}
\end{equation}
and write \(\mathcal{G}_{\tau}=(\mathcal{C},\mathcal{R}_{\tau})\).
Each selected edge is passed to the detector with a propagation weight $w_{AB}$,
\begin{equation}
\label{eq:w_ab}
w_{AB} = \begin{cases}
\max(\Delta(A \to B),\, 0) & \text{if } (A,B) \in \mathcal{R}_V, \\
w_S(A,B) & \text{otherwise,}
\end{cases}
\end{equation}
so validated edges contribute their reduction effect, while pure-observational edges contribute their Suppes geomean. Both quantities lie in $[0,1]$, allowing for a single propagation threshold $\tau_{\mathrm{GI}}$ in Stage~2 (\eqref{eq:r_tau_t}). The resulting graph $\mathcal{G}_\tau$ keeps the validated causal anchor intact and extends it with score-filtered correlation edges. The ablation in §\ref{sec:main_results} also reports $\mathcal{G}_V$ as an alternative option.

\paragraph{\our Trace-Conditioned Graph Guidance.}
The graph from previous steps provides a holistic error dependency pattern across the corpus. \our therefore uses a progressive attribution procedure that activates this pattern per trace to surface plausible downstream errors. In the first stage, the judge receives the trace, taxonomy, and full graph guidance, and returns an initial prediction:
\begin{equation}
\label{eq:e1}
\hat{E}^{(1)}(T) = \mathcal{J}(T,\mathcal{C},\mathcal{R}_\tau).
\end{equation}
\begin{equation}
\label{eq:dt}
D(T) = \{c : \exists\,\ell,\ (c,\ell) \in \hat{E}^{(1)}(T)\}.
\end{equation}
We then derive a trace-specific edge subset by aggregating selected edges from detected source categories onto undetected targets:
\begin{equation}
\label{eq:r_tau_t}
\begin{aligned}
\mathcal{R}_\tau^{(T)} = {} & \bigl\{\,A \!\to\! B \in \mathcal{R}_\tau : \\
 & A\in D(T),\ B\notin D(T), \\
 & \sum_{A' \in D(T)} w_{A'B} > \tau_{\mathrm{GI}}\,\bigr\}.
\end{aligned}
\end{equation}
where $\tau_{\mathrm{GI}}$ is the graph-injection threshold, which controls which downstream hypotheses are exposed in Stage 2. The subset $\mathcal{R}_\tau^{(T)}$ retrieves downstream hypotheses from the fixed detector graph $\mathcal{G}_\tau$ according to the Stage-1 error profile.

When \(\mathcal{R}_\tau^{(T)}\) is non-empty, a second stage issues a targeted prompt that lists the Stage-1 predictions as already detected, presents \(\mathcal{R}_\tau^{(T)}\) as trace-specific guidance, and instructs the judge to output only errors whose category does not appear in \(D(T)\). We denote this Stage-2 prediction by $\hat E^{(2)}(T)$. The final prediction $\hat E^{\our}(T)$ merges Stage~1 with the Stage-2 entries whose categories were not already covered in Stage~1. The fixed graph is not updated at inference time, and only the edge subset exposed in the second stage is conditioned on the trace (details in Appendix~\ref{app:detection_pipeline}).

This two-stage design turns the selected graph from a passive context block into an explicit refinement mechanism. Stage 1 uses the full graph to produce an initial error profile. Stage 2 converts that profile into a restricted set of plausible hypotheses and verifies whether those errors are supported by the trace. For span-level localization benchmarks, we additionally prepend a compact span-identifier index to ground predicted locations. To validate the necessity of Stage 2, we compare against a naive variant that prepends the full graph as static, trace-agnostic context in a single LLM call (the +CG ablation; Table~\ref{tab:cg_main_results}).

\section{Experiments}

\subsection{Experimental Setup}


We evaluate \our on two public benchmarks \textbf{TRAIL}~\cite{deshpande2025trail} and \textbf{MAST} under their original terms. Details about these two benchmarks, eleven open- and close-weight backbones, the compute budget, and the software packages used are listed and cited in Appendix~\ref{app:setup_resources}. Downstream evaluation follows the native prediction target of each benchmark. TRAIL is evaluated as span-level multi-error attribution, requiring both error category and location, whereas MAST is evaluated as trace-level multi-label error classification without step prediction. For MAST, we additionally annotate the first occurrence of each error category solely for event-based graph construction (refer to Appendix~\ref{app:step} for details).

%
%

\begin{table*}[!t]
\centering
\footnotesize
\setlength{\tabcolsep}{3.5pt}
\renewcommand{\arraystretch}{1.0}
\begin{tabular}{l l rrr @{\hspace{6pt}} rrr @{\hspace{6pt}} rrrr}
\toprule
\multirow{2}{*}{\textbf{Model}} &
\multirow{2}{*}{\textbf{Method}} &
\multicolumn{3}{c}{\textbf{TRAIL-GAIA}} &
\multicolumn{3}{c}{\textbf{TRAIL-SWE-Bench}} &
\multicolumn{4}{c}{\textbf{MAST}} \\
\cmidrule(lr){3-5} \cmidrule(lr){6-8} \cmidrule(lr){9-12}
& & F1 & Loc & Joint & F1 & Loc & Joint & F1 & Precision & Recall & Acc \\
\midrule
\multirow{3}{*}{Mistral-Small-3.1-24B}
  & Baseline    & 24.06 & 23.83 &  3.78 &  9.80 &  9.36 &  1.57 & 37.73 & 29.87 & 31.50 & 62.01 \\
  & \our        & 34.15 & 25.71 & 12.27 & 14.40 &  7.59 &  0.87 & 38.29 & 28.55 & 34.59 & 59.86 \\
  & $\Delta$     & \dup{+10.09} & \dup{+1.88} & \dup{+8.49} & \dup{+4.60} & \ddn{-1.77} & \ddn{-0.70} & \dup{+0.56} & \ddn{-1.32} & \dup{+3.09} & \ddn{-2.15} \\[2pt]
\multirow{3}{*}{GPT-oss-120B}
  & Baseline    & 25.76 & 16.92 &  4.59 & 22.08 &  2.58 &  0.00 & 17.84 & 26.28 &  9.47 & 67.02 \\
  & \our        & 37.20 & 28.00 &  9.48 & 30.07 &  1.60 &  0.00 & 27.21 & 29.27 & 18.74 & 64.14 \\
  & $\Delta$     & \dup{+11.44} & \dup{+11.08} & \dup{+4.89} & \dup{+7.99} & \ddn{-0.98} & 0.00 & \dup{+9.37} & \dup{+2.99} & \dup{+9.27} & \ddn{-2.88} \\[2pt]
\multirow{3}{*}{GPT-oss-20B}
  & Baseline    & 23.28 &  6.29 &  2.61 &  9.31 &  0.50 &  0.00 & 17.66 & 26.26 &  9.65 & 67.29 \\
  & \our        & 33.28 & 12.28 &  4.09 & 29.89 &  1.23 &  0.43 & 22.34 & 29.31 & 13.34 & 66.96 \\
  & $\Delta$     & \dup{+10.00} & \dup{+5.99} & \dup{+1.48} & \dup{+20.58} & \dup{+0.73} & \dup{+0.43} & \dup{+4.68} & \dup{+3.05} & \dup{+3.69} & \ddn{-0.33} \\[2pt]
\multirow{3}{*}{Gemma-3-27B-IT}
  & Baseline    & 16.25 &  1.79 &  0.33 & 11.39 &  0.21 &  0.00 & 14.19 & 18.32 &  5.84 & 69.13 \\
  & \our        & 21.26 & 11.20 &  0.73 & 15.41 &  1.47 &  0.00 & 21.17 & 26.97 & 11.37 & 68.10 \\
  & $\Delta$     & \dup{+5.01} & \dup{+9.41} & \dup{+0.40} & \dup{+4.02} & \dup{+1.26} & 0.00 & \dup{+6.98} & \dup{+8.65} & \dup{+5.53} & \ddn{-1.03} \\[2pt]
\multirow{3}{*}{Qwen Family$^{*}$}
  & Baseline    & 12.40 &  3.92 &  0.27 &  2.14 &  0.00 &  0.00 & 16.08 & 28.46 &  9.02 & 66.80 \\
  & \our        & 26.57 & 17.32 &  4.13 & 10.33 &  0.42 &  0.00 & 15.61 & 29.50 &  7.94 & 67.78 \\
  & $\Delta$     & \dup{+14.17} & \dup{+13.40} & \dup{+3.86} & \dup{+8.19} & \dup{+0.42} & 0.00 & \ddn{-0.47} & \dup{+1.04} & \ddn{-1.08} & \dup{+0.98} \\
\midrule
\multirow{3}{*}{Gemini-2.5-Pro}
  & Baseline    & 33.51 & 35.68 & 13.74 & 26.66 &  6.42 &  1.35 & 34.99 & 31.11 & 25.49 & 63.77 \\
  & \our        & 42.13 & 30.15 & 15.26 & 38.97 & 16.58 &  2.01 & 41.57 & 29.04 & 32.97 & 60.79 \\
  & $\Delta$     & \dup{+8.62} & \ddn{-5.53} & \dup{+1.52} & \dup{+12.31} & \dup{+10.16} & \dup{+0.66} & \dup{+6.58} & \ddn{-2.07} & \dup{+7.48} & \ddn{-2.98} \\[2pt]
\multirow{3}{*}{Gemini-2.5-Flash}
  & Baseline    & 37.07 & 34.22 & 12.71 &  8.71 &  0.63 &  0.00 & 24.41 & 26.91 & 16.59 & 64.06 \\
  & \our        & 30.26 & 23.81 &  7.40 & 21.91 &  1.79 &  0.00 & 40.34 & 29.64 & 34.11 & 60.32 \\
  & $\Delta$     & \ddn{-6.81} & \ddn{-10.41} & \ddn{-5.31} & \dup{+13.20} & \dup{+1.16} & 0.00 & \dup{+15.93} & \dup{+2.73} & \dup{+17.52} & \ddn{-3.74} \\[2pt]
\multirow{3}{*}{GPT-4o}
  & Baseline    & 18.73 &  7.45 &  2.45 & 16.72 &  4.19 &  1.33 & 22.87 & 28.00 & 14.78 & 65.32 \\
  & \our        & 28.31 & 10.87 &  1.99 & 26.51 &  1.25 &  0.00 & 29.96 & 31.67 & 20.48 & 64.96 \\
  & $\Delta$     & \dup{+9.58} & \dup{+3.42} & \ddn{-0.46} & \dup{+9.79} & \ddn{-2.94} & \ddn{-1.33} & \dup{+7.09} & \dup{+3.67} & \dup{+5.70} & \ddn{-0.36} \\
\bottomrule
\end{tabular}
\caption{%
  Main results across both benchmarks.
  \textbf{F1} is weighted-F1 on both benchmarks, with the rest macro-averaged.
  \textbf{TRAIL-GAIA} / \textbf{TRAIL-SWE-Bench}: span-level multi-error attribution; \textbf{Loc} location accuracy, \textbf{Joint} location-category joint accuracy.
  \textbf{MAST}: trace-level multi-label yes/no over 13 categories on 393 AG2 traces.
  \textbf{\our}: thresholded inference graph (TRAIL $\tau{=}0.35$; MAST $\tau{=}0.50$) with two-stage dynamic injection.
  \textbf{$\Delta$}: per-cell difference (\our $-$ Baseline).
  All metrics in \%.
  $^{*}$\textbf{Qwen Family} combines the best single-benchmark model per side: QwenLong-L1-32B on TRAIL + QwQ-32B (with \texttt{enable\_thinking}) on MAST.%
}
\label{tab:main_results}
\end{table*}

\paragraph{Graph Construction Protocol.}
We construct one dependency graph per taxonomy from all annotated traces, with categories as nodes and aggregate temporal, probability-raising, and intervention-supported relations as edges, capturing the holistic dependency pattern of the taxonomy. The graph contains no trace-specific labels, spans, evidence, or task answers, and remains fixed during detection. At inference time, the initial error predictions are used to retrieve only downstream hypotheses whose source categories were detected (Stage-1), and a second judge call only verifies those hypotheses against the trace (Stage-2). We therefore interpret the setup as taxonomy-level dependency-guided attribution with controlled context, avoiding trace-specific label leakage.

For each benchmark, we compare baseline against \our with the inference graph $\mathcal{G}_\tau$ (Defined in Section~\ref{graph_guided_detection}). A observational-edge threshold $\tau$ is shared across all backbones within each benchmark and selected by model-averaged weighted F1 over a small candidate grid. This yields $\tau{=}0.35$ for TRAIL and $\tau{=}0.50$ for MAST. Ablations on neighboring thresholds, causal-only graphs, and random controls are in Appendix~\ref{app:results_details}. We report weighted-F1, Loc, and Joint on TRAIL, and weighted-F1, macro-precision, macro-recall, and macro-accuracy on MAST.

\subsection{Main Results}
\label{sec:main_results}




Table~\ref{tab:main_results} evaluates \our on span-level (TRAIL) and trace-level (MAST) multi-error attribution. On TRAIL, \our substantially improves attribution. Gemini-2.5-Pro achieves the strongest results, with $42.13$ F1 on GAIA and $38.97$ on SWE-Bench, while open-weight models show large F1 gains (up to $+21$ on SWE for GPT-oss-20B). Though graph guidance reliably enhances category detection, exact localization on long-context SWE-Bench remains difficult~\cite{deshpande2025trail}. On MAST, graph guidance transfers effectively, lifting F1 for GPT-4o ($22.87\!\to\!29.96$) and all open-weight backbones (up to $+9.4$) except QwQ-32B. However, the accuracy is traded for the gains, with accuracy decreasing on 7 of 8 backbones ($-0.33$ to $-3.74$) while recall rises on the same 7 of 8 ($+3.09$ to $+17.52$). The imbalanced distribution of the 13 binary category decisions per trace, of which only $30.5\%$ are positive, means accuracy is best read alongside recall rather than alone.\footnote{An always-``no'' classifier scores $69.5\%$ accuracy, above every cell in Table~\ref{tab:main_results}.} Results for two newest backbones, GPT-5 and Qwen3.6-35B-A3B, are reported in Appendix~\ref{app:additional_results}, where \our improves F1 in all six cells under the identical configuration. Stage-2 rates in Appendix~\ref{app:results_details}.

\begin{figure}[!t]
  \centering
  \includegraphics[width=\linewidth]{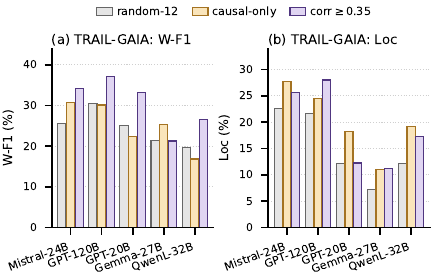}
  \caption{Graph-richness ablation on TRAIL-GAIA across both headline metrics. Three variants: random-12 (edge-count-matched control with no causal screening), causal-only ($\mathcal{G}_V$, intervention-validated edges), and corr-union $\mathcal{G}_\tau$ at $\tau{=}0.35$.}
  \label{fig:threshold_ablation_gaia}
\end{figure}

To confirm that these gains come from the \emph{structure} of the learned graph rather than from prepending any block of edges, we ablate against a random-12 control (matched edge count, no causal screening) alongside the validated-only $\mathcal{G}_V$ and the inference $\mathcal{G}_\tau$. Across the five open-weight backbones in Figure~\ref{fig:threshold_ablation_gaia}, random-12 underperforms the inference graph on 4 of 5 and $\mathcal{G}_\tau$ at $\tau{=}0.35$ takes the F1 peak on 4 of 5 (Gemma-27B is the lone exception, where causal-only wins F1).


\begin{table*}[!tbp]
\centering
\footnotesize
\setlength{\tabcolsep}{3.5pt}
\renewcommand{\arraystretch}{1.0}
\begin{tabular}{l l rrr @{\hspace{6pt}} rrr @{\hspace{6pt}} rrrr}
\toprule
\multirow{2}{*}{\textbf{Model}} &
\multirow{2}{*}{\textbf{Method}} &
\multicolumn{3}{c}{\textbf{TRAIL-GAIA}} &
\multicolumn{3}{c}{\textbf{TRAIL-SWE-Bench}} &
\multicolumn{4}{c}{\textbf{MAST}} \\
\cmidrule(lr){3-5} \cmidrule(lr){6-8} \cmidrule(lr){9-12}
& & F1 & Loc & Joint & F1 & Loc & Joint & F1 & P & R & Acc \\
\midrule
\multirow{3}{*}{Mistral-Small-3.1-24B}
  & +CG      & 30.81 & 27.21 & 12.74 &  7.46 &  1.42 &  0.00 & 37.71 & 28.43 & 30.21 & 63.89 \\
  & \our     & 34.15 & 25.71 & 12.27 & 14.40 &  7.59 &  0.87 & 38.29 & 28.55 & 34.59 & 59.86 \\
  & $\Delta$ & \dup{+3.34} & \ddn{-1.50} & \ddn{-0.47} & \dup{+6.94} & \dup{+6.17} & \dup{+0.87} & \dup{+0.58} & \dup{+0.12} & \dup{+4.38} & \ddn{-4.03} \\[2pt]
\multirow{3}{*}{GPT-oss-120B}
  & +CG      & 28.96 & 17.85 &  6.88 & 23.78 &  0.42 &  0.00 & 21.63 & 28.49 & 12.53 & 66.12 \\
  & \our     & 37.20 & 28.00 &  9.48 & 30.07 &  1.60 &  0.00 & 27.21 & 29.27 & 18.74 & 64.14 \\
  & $\Delta$ & \dup{+8.24} & \dup{+10.15} & \dup{+2.60} & \dup{+6.29} & \dup{+1.18} & 0.00 & \dup{+5.58} & \dup{+0.78} & \dup{+6.21} & \ddn{-1.98} \\[2pt]
\multirow{3}{*}{GPT-oss-20B}
  & +CG      & 12.49 &  7.23 &  0.60 & 12.59 &  0.42 &  0.00 & 18.05 & 27.76 & 10.39 & 67.78 \\
  & \our     & 33.28 & 12.28 &  4.09 & 29.89 &  1.23 &  0.43 & 22.34 & 29.31 & 13.34 & 66.96 \\
  & $\Delta$ & \dup{+20.79} & \dup{+5.05} & \dup{+3.49} & \dup{+17.30} & \dup{+0.81} & \dup{+0.43} & \dup{+4.29} & \dup{+1.55} & \dup{+2.95} & \ddn{-0.82} \\[2pt]
\multirow{3}{*}{Gemma-3-27B-IT}
  & +CG      & 17.58 &  7.63 &  0.20 & 23.39 &  0.36 &  0.00 & 12.17 & 25.26 &  5.39 & 68.72 \\
  & \our     & 21.26 & 11.20 &  0.73 & 15.41 &  1.47 &  0.00 & 21.17 & 26.97 & 11.37 & 68.10 \\
  & $\Delta$ & \dup{+3.68} & \dup{+3.57} & \dup{+0.53} & \ddn{-7.98} & \dup{+1.11} & 0.00 & \dup{+9.00} & \dup{+1.71} & \dup{+5.98} & \ddn{-0.62} \\[2pt]
\multirow{3}{*}{Qwen Family}
  & +CG      & 16.22 &  7.72 &  2.48 &  5.00 &  0.00 &  0.00 & 12.85 & 32.26 &  5.92 & 68.66 \\
  & \our     & 26.57 & 17.32 &  4.13 & 10.33 &  0.42 &  0.00 & 15.61 & 29.50 &  7.94 & 67.78 \\
  & $\Delta$ & \dup{+10.35} & \dup{+9.60} & \dup{+1.65} & \dup{+5.33} & \dup{+0.42} & 0.00 & \dup{+2.76} & \ddn{-2.76} & \dup{+2.02} & \ddn{-0.88} \\
\bottomrule
\end{tabular}
\caption{Single-stage Static Graph Guidance (\textbf{+CG}) vs.\ \our on the five open-weight backbones; $\Delta$ = \our $-$ +CG. +CG prepends the same inference graph ($\tau{=}0.35$ TRAIL, $\tau{=}0.50$ MAST) as a static context block in one call. Columns, metrics, and the no-graph baseline follow Table~\ref{tab:main_results}. All metrics in \%.}
\label{tab:cg_main_results}
\end{table*}

\paragraph{Static Graph Guidance (+CG) Ablation.}
We next test whether the gains require the two-stage consumption of the graph or merely access to it. The +CG ablation prepends the identical edge set $\mathcal{R}_\tau$ as static, trace-agnostic context in a single LLM call (Appendix~\ref{app:detection_pipeline}). Table~\ref{tab:cg_main_results} shows that +CG is inconsistent against the no-graph baseline of Table~\ref{tab:main_results}---helping on some cells (e.g., $+6.8$ F1 on Mistral-3.1 TRAIL-GAIA, $+12.0$ on Gemma-3 TRAIL-SWE-Bench) but degrading on others (e.g., $-10.8$ on GPT-oss-20B TRAIL-GAIA, $-3.2$ on Qwen MAST)---and underperforms \our on 14 of 15 backbone cells (the exception is Gemma-3-27B-IT on TRAIL-SWE-Bench), confirming that the trace-conditioned two-stage design, rather than static graph access, is the operational lever; the mechanism analysis is in Appendix~\ref{app:detection_pipeline}.

%

\begin{table*}[!t]
\centering \footnotesize
\setlength{\tabcolsep}{5pt}
\renewcommand{\arraystretch}{1.0}
\begin{tabular}{l l rrr @{\hspace{10pt}} rrrr}
\toprule
\multirow{2}{*}{\textbf{Model}} &
\multirow{2}{*}{\textbf{Method}} &
\multicolumn{3}{c}{\textbf{TRAIL} (25 held-out traces)} &
\multicolumn{4}{c}{\textbf{MAST} (74 held-out records)} \\
\cmidrule(lr){3-5} \cmidrule(lr){6-9}
& & F1 & Loc & Joint & F1 & Precision & Recall & Acc \\
\midrule
\multirow{3}{*}{Mistral-Small-3.1-24B}
  & Baseline & 24.74 & 27.11 &  3.31 & 34.53 & 26.90 & 27.28 & 58.73 \\
  & \our     & 33.14 & 26.00 &  7.60 & 36.44 & 27.50 & 29.33 & 59.15 \\
  & $\Delta$ & \dup{+8.40} & \ddn{-1.11} & \dup{+4.29} & \dup{+1.91} & \dup{+0.60} & \dup{+2.05} & \dup{+0.42} \\[2pt]
\multirow{3}{*}{GPT-oss-120B}
  & Baseline & 31.08 &  4.67 &  1.60 & 22.09 & 36.73 & 12.87 & 67.46 \\
  & \our     & 33.48 &  9.33 &  4.13 & 34.50 & 39.27 & 22.15 & 66.01 \\
  & $\Delta$ & \dup{+2.40} & \dup{+4.66} & \dup{+2.53} & \dup{+12.41} & \dup{+2.54} & \dup{+9.28} & \ddn{-1.45} \\
\bottomrule
\end{tabular}
\caption{%
  Stratified 80/20 split per benchmark. The detector is scored on the untouched held-out traces. TRAIL combines the GAIA and SWE-Bench splits. All metrics in \%.%
}
\label{tab:holdout_results}
\end{table*}

\paragraph{Held-out Graph Validation.}
To rule out the possibility that the observed gains arise from constructing the graph using evaluation traces, we repeat the pipeline with a stratified 80/20 train/test split for each benchmark. The learned graph is rebuilt from the training side only, so the detector is scored only on the held-out traces. Table~\ref{tab:holdout_results} reports the result: \our improves weighted-F1 in all four cells spanned by two representative backbones, by $+8.40$ and $+2.40$ in TRAIL and $+1.91$ and $+12.41$ in MAST. The graph signal is transferable to traces it never saw, and the improvements in Table~\ref{tab:main_results} are not explained by fitting the evaluation corpus.

\subsection{Comparison with Who\&When-Style Attribution}
\label{sec:ww}


To evaluate if the learned \our graph improves multi-error attribution across different prompting strategies, we adapt Who\&When's \textsc{All-at-once} and \textsc{Step-by-step} regimes, preserving their original output spaces and stopping criteria. Details can be found in Appendix~\ref{app:ww_adaptation}.

\begin{figure}[!htbp]
  \centering
  \includegraphics[width=\linewidth]{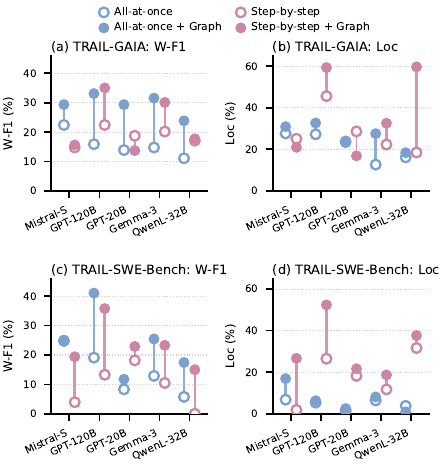}
  \caption{Who\&When-style adaptation on TRAIL across both splits and both headline metrics. Color = strategy (blue \textsc{All-at-once}, pink \textsc{Step-by-step}); slope direction (filled dot above or below open dot) shows whether the graph helps or hurts.}
  \label{fig:ww_bars}
\end{figure}


First, we observe that across the five open-weight backbones, graph guidance consistently benefits both TRAIL splits. \textsc{All-at-once+Graph} improves F1 across nearly all of them (Figure~\ref{fig:ww_bars}). \textsc{Step-by-step+Graph} yields uniform gains on SWE-Bench, boosting larger models such as GPT-oss-120B, while exhibiting minor volatility on GAIA for GPT-oss-20B. Furthermore, a graph-guided variant achieves the best MAST F1 across all model blocks shown in Appendix~\ref{app:ww_adaptation}. \textbf{Overall, the \our graph demonstrates strong transferability across various prompting templates.}

\subsection{Causal Graphs as Explanatory Artifacts}

The preceding results evaluate graph guidance as a prediction prior. Next, we examine the explanatory role of the intervention-validated graph $\mathcal{G}_V$. The inference graph $\mathcal{G}_\tau$ augments $\mathcal{G}_V$ with score-filtered observational edges to improve detection coverage, but those added edges are not all causal. They can recover missing downstream categories, but may also overclaim propagation to categories without causal evidence, and their reliability may not transfer under distribution shift. The validated graph $\mathcal{G}_V$, by contrast, is explanation-oriented. Each edge is supported by controlled counterfactual rollout in which repairing the source error reduces the downstream target. We therefore use $\mathcal{G}_\tau$ for detection and reserve $\mathcal{G}_V$ for explanation, repair prioritization, and propagation analysis.

\begin{figure}[!htbp]
  \centering
  \includegraphics[width=\linewidth]{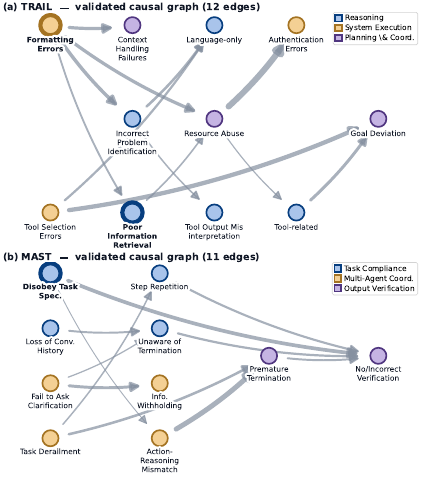}
  \caption{Intervention-validated propagation graphs $\mathcal{G}_V$
    for (a) TRAIL and (b) MAST. Arrows are causal edges supported by
    counterfactual rollout; arrow width $\propto |\Delta(A \to B)|$;
    node color = top-level taxonomy parent; columns = topological
    generations (propagation reads left-to-right). Thick-bordered nodes
    source a propagation chain of length $\ge 3$ edges.}
  \label{fig:gv_graphs}
\end{figure}

Figure~\ref{fig:gv_graphs} visualizes the intervention-validated propagation graph $\mathcal{G}_V$ for TRAIL and MAST, and reveal intervention-backed propagation between error categories rather than mere co-occurrence. For example, a TRAIL edge \textit{Formatting Errors} $\rightarrow$ \textit{Context Handling Failures} indicates that correcting the upstream formatting issue reduced later context-handling failures in counterfactual rollout. Analogously on MAST, \textit{Action-Reasoning Mismatch} $\rightarrow$ \textit{Premature Termination} identifies an upstream repair target before the agent loop ends prematurely. The propagation graphs reveal two main findings. First, the propagation chain can guide agent repair strategy, where fixing upstream nodes with many validated outgoing effects (highlighted in Figure~\ref{fig:gv_graphs}) reduces more downstream errors than repairing the symptoms directly. Second, the graph exposes vulnerable parts of the agent system by surfacing recurring upstream causes. Beyond the predictive contribution of $\mathcal{G}_\tau$, the causal graph $\mathcal{G}_V$ adds a trustworthy and inspectable propagation structure for system-level repair.

\subsection{Error Attribution Analysis}
\label{sec:case_study}

\begin{figure}[!t]
  \centering
  \includegraphics[width=\linewidth]{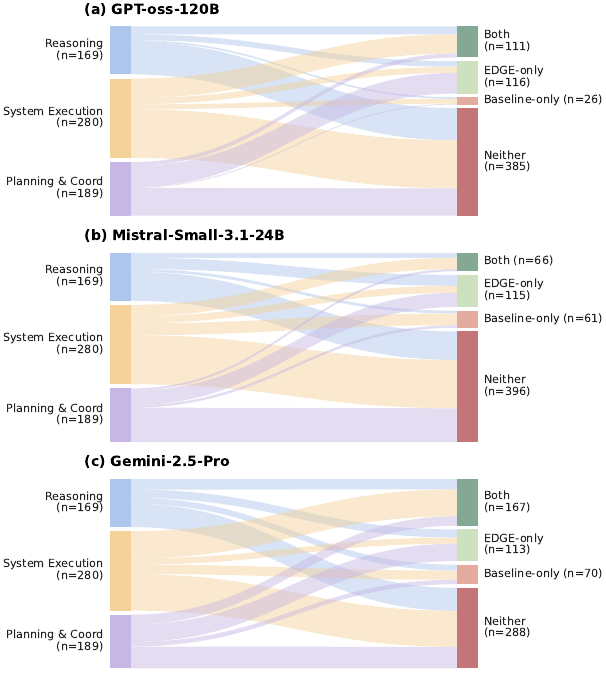}
  \caption{Per-error attribution transitions on TRAIL-GAIA ($n{=}638$ gold-error \emph{instances} flagged by annotators, not traces). Each error flows from its TRAIL parent (left) into one of four buckets defined by which method attributed it: \textbf{Both}, \textbf{+EDGE only}, \textbf{Baseline only}, or \textbf{Neither}. Ribbon width is proportional to count.}
  \label{fig:error_attribution_4bucket}
\end{figure}

To probe whether \our's gains concentrate on the categories its causal graph is designed to surface, we route each of the 638 gold-error instances on the TRAIL-GAIA split into one of four mutually exclusive outcome buckets defined by which method attributed it. We show this transition in Figure~\ref{fig:error_attribution_4bucket} for three backbones spanning open- and closed-source, GPT-oss-120B, Mistral-Small-3.1-24B, and Gemini-2.5-Pro, adopting the parent category in the TRAIL taxonomy as the analysis unit. We aggregate the buckets into recall (gold errors correctly attributed within each parent category) so cross-method gains are directly comparable. On all three backbones, the +EDGE-only bucket is dominated by Planning \& Coord., yielding the largest recall jump (GPT-oss-120B $9.0\% {\to} 46.6\%$, Mistral $7.4\% {\to} 31.2\%$, Gemini-2.5-Pro $26.5\% {\to} 50.8\%$), with Reasoning second. System Execution barely moves on GPT-oss-120B and drops on Mistral and Gemini-2.5-Pro. This pattern aligns with the topology of $\mathcal{G}_V$, where Planning errors are downstream targets recovered by Stage~2, Reasoning errors are intermediate nodes with moderate gains, and System errors are upstream sources whose local cues benefit less from graph guidance. Detailed examples are in Appendix~\ref{app:case_study}.

\FloatBarrier
\section{Conclusion}

We presented \our, a graph-guided framework for multi-error attribution in LLM agent traces that constructs and exploits an error dependency graph. \our is taxonomy-agnostic, so the construction and detection pipeline can be rerun for any error taxonomy with observed error events. Experiments on TRAIL and MAST show that the inference graph improves category-level multi-error attribution across both span-level and trace-level settings. These results show that error dependencies provide useful structure beyond isolated label prediction. By exposing how local failures propagate, \our supports more explainable debugging, upstream repair prioritization, and more reliable improvement of long-horizon agent systems.

%
%
%
%
%
%

\section*{Limitations}

Our evaluation is best interpreted as taxonomy-level dependency-guided attribution. Because most agent-error annotation identifies only a single root error, corpora with multi-error annotations remain scarce and have low per-category support, so the dependency graph is constructed from the full annotated corpus and remains fixed during detection, containing only aggregate category-level dependencies rather than trace-specific labels, spans, or task answers. Repair verification, effect evaluation, and MAST event annotation rely on LLM judgments, which may introduce noise. Because the graph is estimated from category co-occurrence, it helps more with detecting which error events occur than with localizing them along the trajectory, and the localization gains are correspondingly smaller, staying close to flat on the long-context SWE-Bench split.

Our intervention validation holds tool outputs fixed, so each estimated edge effect reflects error propagation along the agent's reasoning path rather than mediation through downstream tool re-execution. Intervention validation is also limited by the number of co-occurring traces per candidate edge, so we extend the validated causal subset with additional correlation-supported edges to maintain coverage. The validated causal subset alone serves as the more reliable structure for explanation and repair analysis.

The MAST event annotations are automatically produced and not fully human-validated. Coverage is reported but does not measure event-position correctness, and noisy positions may affect temporal-priority estimates and the inferred graph. Fully human-validated event annotations would strengthen future evaluation.

\our applies graph guidance uniformly to every trace, and gating the injection on model confidence or task complexity is a natural refinement for the cases where guidance does not help. Beyond diagnosis, the intervention-validated subgraph $\mathcal{G}_V$ is the artifact a repair loop would consume, and evaluating such a loop faithfully requires re-executing multi-agent systems at scale, which we leave to future work.

\paragraph{Potential Risks.}
\our could be misused if graph-guided outputs are treated as definitive causal diagnoses. \our should be used as a diagnostic aid with human review, especially in high-stakes agent deployments.

\paragraph{Reproducibility.} All reported numbers are single-run point estimates. The full code and inference scripts are available at \url{https://github.com/JuneHou/EDGE}. However, we do not guarantee that others will obtain exactly the same numbers as those reported in our paper, due to the inherent nondeterminism in LLM inference.

\section*{Ethical Considerations}
This study evaluates multi-agent failure attribution using publicly available benchmarks under their original terms. We do not collect new human-subject data or use private user data. Some parts of the pipeline rely on LLM judgments, including repair verification, effect evaluation, and MAST event annotation, which may introduce annotation noise or model bias. We report these limitations and encourage future work with stronger human validation.

\paragraph{AI Assistance.} We used AI assistants for parts of the implementation and manuscript preparation, including generating LaTeX code for tables and refining text written by the authors. All AI-generated content was carefully reviewed and revised by the authors to ensure accuracy and clarity.

\section*{Acknowledgments}
This research is sponsored by NSF 2442253, 2607580, NIH 1R21AG091260-01, USDA NIFA, Commonwealth Cyber Initiative, and generous gifts from Nvidia, Cisco, and the Amazon-Virginia Tech Initiative. This research used the Delta system at the National Center for Supercomputing Applications [award OAC 2005572] through allocation [NAIRR240202] from the Advanced Cyberinfrastructure Coordination Ecosystem: Services \& Support (ACCESS) program, which is supported by National Science Foundation grants \#2138259, \#2138286, \#2138307, \#2137603, and \#2138296.

\bibliography{custom}

\appendix

\section{Appendix}
\label{sec:appendix}

\subsection{Notation Table} \label{app:notation}

Table~\ref{tab:graph-notation} lists the symbols used in \our. We describe the role of each symbol here in prose, grouped by the three blocks shown in the table.

\paragraph{Graphs and edge sets.} $\mathcal{G}_S$ is the observational correlation graph from the Suppes screen. Its edges are the observational hypotheses that feed coverage augmentation in $\mathcal{G}_\tau$. $\mathcal{G}_C$ is the CAPRI-pruned DAG of causal candidates fed into intervention validation. $\mathcal{G}_V$ is the intervention-validated causal anchor used for explanation. $\mathcal{G}_\tau$ is the prediction-oriented thresholded inference graph that unions $\mathcal{G}_V$ with score-filtered observational edges from $\mathcal{G}_S$, and is the default detector input.

\paragraph{Sets and trace-level quantities.} $\mathcal{C}$ is the error taxonomy (nodes of all graphs). $\mathcal{T}$ is the trace corpus. $\mathcal{T}_{A\prec B}$ is the subset of traces in which both $A$ and $B$ occur with $t_A(T)<t_B(T)$. $X_A(T)\in\{0,1\}$ indicates whether $A$ occurs in $T$. $t_A(T)$ is the index of the first span at which $A$ emerges.

\paragraph{Edge scores and thresholds.} $w_S(A,B)$ is the Suppes propagation weight (geomean of temporal precedence and probability raising). $\Delta(A\!\to\!B)$ is the controlled direct effect (Boolean risk-difference of $B$ between the original trace and the controlled-counterfactual rollout). $\tau$ is the inference-graph threshold, the minimum $w_S$ for an observational edge to enter $\mathcal{G}_\tau$. $\tau_{\mathrm{GI}}$ is the Stage-2 propagation threshold, the minimum aggregated $w_{AB}$ for a downstream hypothesis to be exposed in Stage~2. $\tau_{\mathrm{val}}$ is the edge-validation threshold, the value that $\Delta(A\!\to\!B)$ must exceed for a causal-candidate edge to enter $\mathcal{G}_V$.

\begin{table}[!tbp]
\centering
\small
\setlength{\tabcolsep}{6pt}
\renewcommand{\arraystretch}{1.15}
\begin{tabular}{ll}
\toprule
Symbol & Name \\
\midrule
$\mathcal{G}_S=(\mathcal{C}, \mathcal{R}_S)$ & Observational correlation graph \\
$\mathcal{G}_C=(\mathcal{C}, \mathcal{R}_C)$ & Causal-candidate graph \\
$\mathcal{G}_V=(\mathcal{C}, \mathcal{R}_V)$ & Validated graph \\
$\mathcal{G}_\tau=(\mathcal{C}, \mathcal{R}_\tau)$ & Thresholded inference graph \\
\midrule
$\mathcal{C}$ & Error taxonomy \\
$\mathcal{T}$ & Trace corpus \\
$\mathcal{T}_{A\prec B}$ & Joint-event subset \\
$X_A(T)\in\{0,1\}$ & Category indicator \\
$t_A(T)$ & First-event position \\
\midrule
$w_S(A,B)$ & Suppes propagation weight \\
$\Delta(A \to B)$ & Controlled direct effect \\
$\tau$ & Inference-graph threshold \\
$\tau_{\mathrm{GI}}$ & Stage-2 propagation threshold \\
$\tau_{\mathrm{val}}$ & Edge-validation threshold \\
\bottomrule
\end{tabular}
\caption{Notation used in \our (symbol roles described in text).}
\label{tab:graph-notation}
\end{table}

\subsection{Causal Graph Construction Details} \label{app:graph_construction}

This appendix collects the formal definitions, threshold values, prompts, postcheck criteria, and hyperparameter values used for the observational correlation graph $\mathcal{G}_S$, the causal-candidate graph $\mathcal{G}_C$ constructed in Section~\ref{candidate_graph}, and the intervention-based validation pipeline of Section~\ref{graph_validation}.

\paragraph{Adapting Suppes screening to within-trace event data.}
Classic CAPRI often uses marginal occurrence frequency as a temporal-priority surrogate when only cross-sectional event observations are available. Our setting differs because each trace carries span-level event annotations, so we can estimate temporal priority directly from within-trace event order and aggregate this evidence across traces.

\paragraph{First-event formalization.}
For each annotated trace $T$ and category $A$ with $X_A(T)=1$, the first-event index $t_A(T)$ is the position of the first span at which $A$ emerges:
\[
\begin{aligned}
t_A(T) = \min\bigl\{\,i :\ & \exists(c,\ell)\in E(T), \\
 & c=A,\ \ell=s_i\,\bigr\}.
\end{aligned}
\]
Tied events are retained as category occurrences but do not support strict temporal direction.

\paragraph{CAPRI score objective.}
We select the pruned graph by minimizing an AIC-style score over the admissible family $\Omega(\mathcal{R}_S)$ defined in the main text:
\[
\begin{aligned}
\mathcal{G}_C = \arg\min_{\mathcal{G}\in\Omega(\mathcal{R}_S)} \bigl[ & -2\log \widehat{L}(\mathcal{G};X) \\
 & + 2\,\mathrm{dim}(\mathcal{G})\bigr],
\end{aligned}
\]
where $\widehat{L}(\mathcal{G};X)$ is the maximized likelihood of the binary category-event model and $\mathrm{dim}(\mathcal{G})$ is the number of free parameters. Optimization proceeds by hill-climbing over edge addition, removal, and reversal, restricted to $\mathcal{R}_S$ and enforcing acyclicity.

\paragraph{Graph construction yields.}
Table~\ref{tab:graph_construction_yields} reports the edge yield at each construction stage for both benchmarks under the production AIC configuration (\texttt{min\_precedence}=0.55, \texttt{min\_pr\_delta}=0.05, \texttt{min\_joint}=3). MAST starts from a larger ordered-pair pool than TRAIL because its taxonomy is flat (13 codes vs.\ 19 hierarchical leaves) and its $393$ trace corpus produces more co-occurrence support per category. The causal-candidate graph $\mathcal{G}_C$ therefore retains a larger DAG (23 vs.\ 13 edges) before intervention validation.

\begin{table}[!tbp]
\centering
\footnotesize
\setlength{\tabcolsep}{4pt}
\renewcommand{\arraystretch}{1.05}
\begin{tabular}{l rr}
\toprule
\textbf{Stage} & \textbf{TRAIL} & \textbf{MAST} \\
\midrule
Traces                                  & 148   & 393   \\
Categories                              &  19   &  13   \\
Ordered pairs                           &  --   & 1{,}347 \\
\midrule
$|\mathcal{R}_S|$ (Suppes)              &  27   &  43   \\
$|\mathcal{R}_C|$ (CAPRI)               &  13   &  23   \\
\quad hill-climb moves accepted         &  13   &  25   \\
\midrule
Bootstrap + shuffle controls            & skipped & run \\
\bottomrule
\end{tabular}
\caption{Edge yield at each graph-construction stage for both benchmarks (production AIC configuration). The Suppes screen produces $\mathcal{R}_S$, and CAPRI hill-climb prunes it to $\mathcal{R}_C$. TRAIL skips bootstrap and shuffle controls due to small $N$. MAST runs both ($n_{\mathrm{bootstrap}}{=}100$, $n_{\mathrm{shuffles}}{=}50$).}
\label{tab:graph_construction_yields}
\end{table}

\paragraph{Intervention trace set.}
For a causal-candidate edge $A\rightarrow B \in \mathcal{R}_C$, the set of traces eligible for intervention is
\[
\begin{aligned}
\mathcal{T}_{A\prec B} = \bigl\{\,T\in\mathcal{T}: & \ X_A(T){=}1,\ X_B(T){=}1, \\
 & \ t_A(T)<t_B(T)\,\bigr\}.
\end{aligned}
\]
This restricts intervention experiments to traces in which both categories occur and the first event of $A$ strictly precedes the first event of $B$, ensuring that any change in $B$'s downstream status can be attributed to the intervention on $A$ rather than to baseline ordering.

\paragraph{Per-category patch library.}
The patch library specifies, for each source category $A$, what a valid \(do(A{=}0)\) repair must look like. Each entry contains: \texttt{category} and \texttt{trail\_definition} (the taxonomy reference), a \texttt{patch\_side\_default} that fixes whether the repair rewrites the source span's input context or its output, a \texttt{slot\_schema} that names the structured slots the repair LLM must fill, a natural-language \texttt{repair\_instruction}, a list of \texttt{forbidden\_actions}, and a list of declarative \texttt{postcheck} criteria. TRAIL covers the 19 source-error categories that appear in the dataset under its taxonomy, with two intervention sides (\texttt{replace\_span\_output} for LLM spans, \texttt{replace\_span\_input} for context-side patches). MAST covers the 13 categories that appear under its taxonomy with a single \texttt{replace\_step\_content} mode over multi-agent conversation steps. The full libraries are reproduced below as JSON.

\begin{codebox}{TRAIL patch library (\texttt{causal/patch/patch\_library.json})}
{
  "Formatting Errors": {
    "category": "Formatting Errors",
    "trail_definition": "Formatting / code execution / required structure errors.",
    "patch_side_default": "replace_span_output",
    "slot_schema": {
      "REQUIRED_MARKERS": "List of literal tags or markers explicitly required in output (e.g. <end_plan>)",
      "REQUIRED_FORMAT_RULES": "Explicit structure rules (e.g. comma-separated list, JSON, code fence, FINAL ANSWER: prefix)",
      "FORBIDDEN_MARKERS": "Any explicitly prohibited markers, if stated"
    },
    "error_type_spec_text": "error_type: Formatting Errors\ntrail_definition: Formatting / code execution / required structure errors.\npatch_side_default: replace_span_output\nslot_extraction:\n- REQUIRED_MARKERS: extract any explicit required tags or literal markers from ERROR_DESCRIPTION, ERROR_EVIDENCE, USER_REQUIREMENTS, or LOCAL_SNIPPET.\n- REQUIRED_FORMAT_RULES: extract explicit structure requirements (e.g., exact suffix, JSON, comma-separated list, code fence).\n- FORBIDDEN_MARKERS: extract any explicitly prohibited markers if present.\nrepair_instruction:\n- Make the smallest possible structural edit that satisfies the required format contract.\n- Do not change semantic content unless needed to satisfy the explicit format rule.\n- If a required literal marker is explicitly grounded, append or insert it exactly.\nforbidden_actions:\n- Do not add new factual content.\n- Do not add tool calls, evidence, or reasoning that was not already present.\n- Do not change plan steps unless the only needed change is the structural marker.\npostcheck:\n- All grounded REQUIRED_MARKERS appear exactly.\n- No ungrounded special tokens were introduced.\n- Original semantic content is preserved except for the minimum formatting fix.",
    "repair_instruction": "Make the smallest possible structural edit that satisfies the required format contract. Do not change semantic content unless needed to satisfy the explicit format rule. If a required literal marker is explicitly grounded, append or insert it exactly.",
    "forbidden_actions": [
      "Do not add new factual content.",
      "Do not add tool calls, evidence, or reasoning that was not already present.",
      "Do not change plan steps unless the only needed change is the structural marker."
    ],
    "postcheck": [
      "All grounded REQUIRED_MARKERS appear exactly in patch_payload.",
      "No ungrounded special tokens (e.g. <...>) introduced.",
      "Original semantic content preserved except for minimum formatting fix.",
      "patch_payload differs from local_snippet."
    ]
  },
  "Incorrect Problem Identification": {
    "category": "Incorrect Problem Identification",
    "trail_definition": "Misunderstood the overall task or the local task.",
    "patch_side_default": "replace_span_input",
    "slot_schema": {
      "TASK_GOAL": "Explicit user objective from USER_REQUIREMENTS or task text",
      "REQUIRED_OUTPUT_FORM": "Any answer-format requirement",
      "MISIDENTIFIED_TARGET": "What the current span incorrectly focuses on, if stated in ERROR_DESCRIPTION"
    },
    "error_type_spec_text": "error_type: Incorrect Problem Identification\ntrail_definition: Misunderstood the overall task or the local task.\npatch_side_default: replace_span_input\nslot_extraction:\n- TASK_GOAL: extract the explicit user objective from USER_REQUIREMENTS or task text.\n- REQUIRED_OUTPUT_FORM: extract any answer-format requirement.\n- MISIDENTIFIED_TARGET: extract what the current span incorrectly focuses on, if stated in ERROR_DESCRIPTION.\nrepair_instruction:\n- Rewrite the local instruction/thought so it correctly states the actual task objective and immediate subgoal.\n- Narrow the local scope to what must be solved next.\n- Preserve any useful constraints already present.\nforbidden_actions:\n- Do not fabricate facts or tool outputs.\n- Do not pre-answer the task.\n- Do not insert downstream recovery instructions for other error types.\npostcheck:\n- The rewritten span explicitly reflects the correct task objective.\n- The required output form remains preserved if grounded.\n- No unsupported factual claims were added.",
    "repair_instruction": "Rewrite the local instruction/thought so it correctly states the actual task objective and immediate subgoal. Narrow the local scope to what must be solved next. Preserve any useful constraints already present.",
    "forbidden_actions": [
      "Do not fabricate facts or tool outputs.",
      "Do not pre-answer the task.",
      "Do not insert downstream recovery instructions for other error types."
    ],
    "postcheck": [
      "The rewritten span explicitly reflects the correct task objective.",
      "The required output form remains preserved if grounded.",
      "No unsupported factual claims were added.",
      "patch_payload differs from local_snippet."
    ]
  },
  "Poor Information Retrieval": {
    "category": "Poor Information Retrieval",
    "trail_definition": "Tried to find information not relevant to the task.",
    "patch_side_default": "replace_span_input",
    "slot_schema": {
      "TASK_ENTITY": "Core entity, resource, or fact the task actually needs",
      "TIME_SCOPE": "Any explicit time restriction",
      "SOURCE_CONSTRAINT": "Any explicitly required source (e.g. USGS, file, website)",
      "BAD_RETRIEVAL_FOCUS": "The irrelevant focus described in ERROR_DESCRIPTION, if present"
    },
    "error_type_spec_text": "error_type: Poor Information Retrieval\ntrail_definition: Tried to find information not relevant to the task.\npatch_side_default: replace_span_input\nslot_extraction:\n- TASK_ENTITY: extract the core entity, resource, or fact the task actually needs.\n- TIME_SCOPE: extract any explicit time restriction.\n- SOURCE_CONSTRAINT: extract any explicit required source (e.g., USGS, file, website).\n- BAD_RETRIEVAL_FOCUS: extract the irrelevant focus described in ERROR_DESCRIPTION, if present.\nrepair_instruction:\n- Rewrite the local query/request/plan step so it targets only task-relevant information.\n- Preserve the required source and time scope if explicitly given.\n- Keep the retrieval action narrow and directly aligned to the task.\nforbidden_actions:\n- Do not inject the answer.\n- Do not fabricate retrieved evidence.\n- Do not add extra exploratory searches unrelated to the stated goal.\npostcheck:\n- The revised span is directly relevant to TASK_ENTITY.\n- Explicit source/time constraints are retained when grounded.\n- No unsupported answer content was introduced.",
    "repair_instruction": "Rewrite the local query/request/plan step so it targets only task-relevant information. Preserve the required source and time scope if explicitly given. Keep the retrieval action narrow and directly aligned to the task.",
    "forbidden_actions": [
      "Do not inject the answer.",
      "Do not fabricate retrieved evidence.",
      "Do not add extra exploratory searches unrelated to the stated goal."
    ],
    "postcheck": [
      "The revised span is directly relevant to TASK_ENTITY.",
      "Explicit source/time constraints are retained when grounded.",
      "No unsupported answer content was introduced.",
      "patch_payload differs from local_snippet."
    ]
  },
  "Resource Abuse": {
    "category": "Resource Abuse",
    "trail_definition": "Excessive tool calling due to memory issues / repeated unnecessary use of resources.",
    "patch_side_default": "replace_span_input",
    "slot_schema": {
      "CURRENT_SUBGOAL": "Immediate goal of the step",
      "REPEATED_ACTION": "Redundant action or repeated call pattern from ERROR_DESCRIPTION/EVIDENCE",
      "AVAILABLE_RESULT": "Any already-known result in local context, if explicitly present",
      "STOP_CRITERION": "Minimal completion condition inferred from USER_REQUIREMENTS"
    },
    "error_type_spec_text": "error_type: Resource Abuse\ntrail_definition: Excessive tool calling due to memory issues / repeated unnecessary use of resources.\npatch_side_default: replace_span_input\nslot_extraction:\n- CURRENT_SUBGOAL: extract the immediate goal of the step.\n- REPEATED_ACTION: extract the redundant action or repeated call pattern from ERROR_DESCRIPTION/EVIDENCE.\n- AVAILABLE_RESULT: extract any already-known result in local context, if explicitly present.\n- STOP_CRITERION: infer the minimal completion condition from USER_REQUIREMENTS.\nrepair_instruction:\n- Rewrite the local step to avoid repeating already-completed or redundant actions.\n- Prefer reusing already available local results if explicitly present.\n- Add a concise stop condition or 'do not repeat identical call' constraint in the local wording.\nforbidden_actions:\n- Do not invent successful prior results unless explicitly present.\n- Do not change the overall plan beyond preventing the local redundant action.\n- Do not introduce credentials, auth logic, or other downstream fixes.\npostcheck:\n- The revised span removes the redundant or excessive action.\n- A concrete local stop condition is present.\n- No new unsupported results were added.\noutput_side_note (applies when LOCAL_SNIPPET is a tool_calls JSON object):\n- This occurs when the annotated span was a TOOL span remapped to its parent LLM span.\n  LOCAL_SNIPPET will be the LLM's tool_calls output \u2014 a JSON object like:\n  {\"role\": \"assistant\", \"tool_calls\": [{\"id\": ..., \"function\": {\"name\": ..., \"arguments\": {...}}, \"type\": \"function\"}]}\n- Fix the repetition by modifying the tool call ARGUMENTS (e.g., make the query more\n  specific, remove a redundant filter parameter like filter_year, or change the search\n  term to avoid repeating a prior identical call).\n- You may also replace the entire tool call with a more targeted one.\n- CRITICAL: Do NOT add new top-level fields to tool_call entries (e.g., stop_criterion,\n  stop_condition, retry_limit). The only valid keys per entry are: id, type, function.\n  Adding phantom fields is not valid OpenAI tool_calls schema and WILL fail postcheck.",
    "repair_instruction": "Rewrite the local step to avoid repeating already-completed or redundant actions. Prefer reusing already available local results if explicitly present. Add a concise stop condition or 'do not repeat identical call' constraint in the local wording.",
    "forbidden_actions": [
      "Do not invent successful prior results unless explicitly present.",
      "Do not change the overall plan beyond preventing the local redundant action.",
      "Do not introduce credentials, auth logic, or other downstream fixes."
    ],
    "postcheck": [
      "The revised span removes the redundant or excessive action.",
      "A concrete local stop condition is present.",
      "No new unsupported results were added.",
      "patch_payload differs from local_snippet."
    ]
  },
  "Task Orchestration": {
    "category": "Task Orchestration",
    "trail_definition": "Subtask coordination and progress monitoring failures.",
    "patch_side_default": "replace_span_input",
    "slot_schema": {
      "PLAN_STEPS": "Explicit plan or pending subtasks from local context if present",
      "CURRENT_STEP": "Immediate intended step",
      "SKIPPED_STEP": "Any skipped prerequisite mentioned in ERROR_DESCRIPTION/EVIDENCE",
      "HANDOFF_TARGET": "Required next tool/agent handoff if explicitly grounded"
    },
    "error_type_spec_text": "error_type: Task Orchestration\ntrail_definition: Subtask coordination and progress monitoring failures.\npatch_side_default: replace_span_input\nslot_extraction:\n- PLAN_STEPS: extract the explicit plan or pending subtasks from local context if present.\n- CURRENT_STEP: identify the immediate intended step.\n- SKIPPED_STEP: extract any skipped prerequisite mentioned in ERROR_DESCRIPTION/EVIDENCE.\n- HANDOFF_TARGET: extract any required next tool/agent handoff if explicitly grounded.\nrepair_instruction:\n- Rewrite the local span so the next action follows the correct prerequisite order.\n- Restore any clearly skipped prerequisite step before proceeding.\n- Keep the patch local: only repair the immediate orchestration mistake.\nforbidden_actions:\n- Do not complete multiple future subtasks inside this one patch.\n- Do not fabricate outputs for skipped steps.\n- Do not directly resolve downstream context issues beyond restoring local order.\npostcheck:\n- The revised span executes or requests the correct next prerequisite.\n- No skipped step remains bypassed locally.\n- No fabricated completion of later subtasks is introduced.",
    "repair_instruction": "Rewrite the local span so the next action follows the correct prerequisite order. Restore any clearly skipped prerequisite step before proceeding. Keep the patch local: only repair the immediate orchestration mistake.",
    "forbidden_actions": [
      "Do not complete multiple future subtasks inside this one patch.",
      "Do not fabricate outputs for skipped steps.",
      "Do not directly resolve downstream context issues beyond restoring local order."
    ],
    "postcheck": [
      "The revised span executes or requests the correct next prerequisite.",
      "No skipped step remains bypassed locally.",
      "No fabricated completion of later subtasks is introduced.",
      "patch_payload differs from local_snippet."
    ]
  },
  "Tool Selection Errors": {
    "category": "Tool Selection Errors",
    "trail_definition": "Used the wrong tool for the task.",
    "patch_side_default": "replace_span_input",
    "slot_schema": {
      "TASK_NEED": "What capability is actually required (search, parse, inspect, compute, final answer)",
      "WRONG_TOOL": "The incorrect tool named in the local span if present",
      "AVAILABLE_TOOL_LIST": "Tools explicitly available in local context",
      "CORRECT_TOOL_HINT": "Best supported tool only if justified by TASK_NEED and AVAILABLE_TOOL_LIST"
    },
    "error_type_spec_text": "error_type: Tool Selection Errors\ntrail_definition: Used the wrong tool for the task.\npatch_side_default: replace_span_input\nslot_extraction:\n- TASK_NEED: extract what capability is actually required (search, parse, inspect, compute, final answer).\n- WRONG_TOOL: extract the incorrect tool named in the local span if present.\n- AVAILABLE_TOOL_LIST: extract the tools explicitly available in local context.\n- CORRECT_TOOL_HINT: select the best supported tool only if justified by TASK_NEED and AVAILABLE_TOOL_LIST.\nrepair_instruction:\n- Rewrite the input context so it explicitly directs the agent to use CORRECT_TOOL_HINT.\n- If the agent answered from memory without calling any tool (no tool call in the input's assistant turn), append a short mandatory instruction at the end of the last user or system message: e.g. 'IMPORTANT: You MUST verify this using [CORRECT_TOOL_HINT] before providing a final answer. Do not answer from internal knowledge alone.'\n- If the agent called the wrong tool, replace or annotate the relevant step to redirect to CORRECT_TOOL_HINT instead.\n- Keep the rest of the input unchanged. Only add or modify the minimal context needed to enforce correct tool use.\nforbidden_actions:\n- Do not fabricate tool outputs.\n- Do not add extra tools not listed in local context.\n- Do not answer the task directly unless the original local step was already the final-answer step and no tool is needed.\n- Do not restructure or summarize the entire input \u2014 only add the minimal directive.\npostcheck:\n- The patched input contains an explicit reference to CORRECT_TOOL_HINT.\n- Any tool mentioned is grounded in AVAILABLE_TOOL_LIST.\n- The local subgoal remains the same.",
    "repair_instruction": "Rewrite the input context so it explicitly directs the agent to use CORRECT_TOOL_HINT. If the agent answered from memory, append a short mandatory instruction: 'IMPORTANT: You MUST verify this using [CORRECT_TOOL_HINT] before providing a final answer.' Keep all other input unchanged.",
    "forbidden_actions": [
      "Do not fabricate tool outputs.",
      "Do not add extra tools not listed in local context.",
      "Do not answer the task directly unless the original local step was already the final-answer step and no tool is needed.",
      "Do not restructure or summarize the entire input \u2014 only add the minimal directive."
    ],
    "postcheck": [
      "The patched input contains an explicit reference to CORRECT_TOOL_HINT.",
      "Any tool mentioned is grounded in AVAILABLE_TOOL_LIST.",
      "The local subgoal remains the same.",
      "patch_payload differs from local_snippet."
    ]
  },
  "Instruction Non-compliance": {
    "category": "Instruction Non-compliance",
    "trail_definition": "Failed to perform the task provided and instead did something else / violated explicit instructions.",
    "patch_side_default": "replace_span_output",
    "slot_schema": {
      "MUST_DO": "Explicit required actions from USER_REQUIREMENTS or ERROR_DESCRIPTION",
      "MUST_NOT_DO": "Explicit prohibitions from USER_REQUIREMENTS or ERROR_DESCRIPTION",
      "REQUIRED_OUTPUT_FORM": "Any answer-format requirement",
      "MISSING_STEP": "Concrete missing action implied by the error evidence (e.g., use tool, include citations, follow template)"
    },
    "error_type_spec_text": "error_type: Instruction Non-compliance\ntrail_definition: Failed to perform the task provided and instead did something else / violated explicit instructions.\npatch_side_default: replace_span_output\nslot_extraction:\n- MUST_DO: extract explicit required actions from USER_REQUIREMENTS or ERROR_DESCRIPTION.\n- MUST_NOT_DO: extract explicit prohibitions from USER_REQUIREMENTS or ERROR_DESCRIPTION.\n- REQUIRED_OUTPUT_FORM: extract any answer-format requirement.\n- MISSING_STEP: extract the concrete missing action implied by ERROR_EVIDENCE.\nrepair_instruction:\n- Rewrite the local output so it directly satisfies MUST_DO and REQUIRED_OUTPUT_FORM.\n- Remove any content that violates MUST_NOT_DO.\n- Keep edits minimal and local.\nforbidden_actions:\n- Do not fabricate facts or tool outputs.\n- Do not add new constraints not grounded in inputs.\n- Do not fix downstream errors explicitly.\npostcheck:\n- Output satisfies grounded MUST_DO and REQUIRED_OUTPUT_FORM.\n- No grounded MUST_NOT_DO violations remain.\n- No new unsupported claims were introduced.",
    "repair_instruction": "Rewrite the local output so it directly satisfies MUST_DO and REQUIRED_OUTPUT_FORM, and remove content violating MUST_NOT_DO. Keep edits minimal and local.",
    "forbidden_actions": [
      "Do not fabricate facts or tool outputs.",
      "Do not add new constraints not grounded in inputs.",
      "Do not fix downstream errors explicitly."
    ],
    "postcheck": [
      "Output satisfies grounded MUST_DO and REQUIRED_OUTPUT_FORM.",
      "No grounded MUST_NOT_DO violations remain.",
      "No new unsupported claims were introduced.",
      "patch_payload differs from local_snippet."
    ]
  },
  "Tool Output Misinterpretation": {
    "category": "Tool Output Misinterpretation",
    "trail_definition": "Made assumptions about tool output or used it in an incorrect context.",
    "patch_side_default": "replace_span_input",
    "slot_schema": {
      "TOOL_NAME": "Tool referenced in the local span or context",
      "RAW_OBSERVATION": "The actual tool observation string if present in local context",
      "REQUIRED_FIELDS": "Fields that must be extracted/checked from tool output (if specified)",
      "PARSE_RULE": "How to parse/interpret output (JSON parse, regex, key lookup) based on grounded hints",
      "FAILSAFE_ACTION": "What to do if parsing/fields missing (retry once, ask clarifying question, stop)"
    },
    "error_type_spec_text": "error_type: Tool Output Misinterpretation\ntrail_definition: Made assumptions about tool output or used it in an incorrect context.\npatch_side_default: replace_span_input\nslot_extraction:\n- TOOL_NAME: extract from LOCAL_SNIPPET or USER_REQUIREMENTS if present.\n- RAW_OBSERVATION: extract any explicit observation content present in local context.\n- REQUIRED_FIELDS: extract required fields/checks if specified.\n- PARSE_RULE: derive from grounded format hints (JSON/text/table) only.\n- FAILSAFE_ACTION: set a conservative fallback (retry once or ask clarification).\nrepair_instruction:\n- Rewrite the local instruction/thought to include an explicit 'parse \u2192 validate \u2192 use' step.\n- Require that any claim derived from the tool output must cite the RAW_OBSERVATION content.\n- If validation fails, execute FAILSAFE_ACTION instead of guessing.\nforbidden_actions:\n- Do not fabricate tool outputs.\n- Do not introduce new fields into tool_calls schema.\n- Do not claim a parsed value unless it is present in RAW_OBSERVATION.\npostcheck:\n- Contains explicit parse + validation gate.\n- Contains a failsafe branch.\n- No new unsupported tool-derived claims were added.",
    "repair_instruction": "Rewrite the local instruction/thought to include an explicit parse\u2192validate\u2192use step for tool output. If validation fails, perform FAILSAFE_ACTION instead of guessing.",
    "forbidden_actions": [
      "Do not fabricate tool outputs.",
      "Do not introduce new fields into tool_calls schema.",
      "Do not claim a parsed value unless it is present in RAW_OBSERVATION."
    ],
    "postcheck": [
      "Contains explicit parse + validation gate.",
      "Contains a failsafe branch.",
      "No new unsupported tool-derived claims were added.",
      "patch_payload differs from local_snippet."
    ]
  },
  "Tool-related": {
    "category": "Tool-related",
    "trail_definition": "Fabricated the outcome of a tool interaction or data retrieval step / fabricated tool capabilities.",
    "patch_side_default": "replace_span_output",
    "slot_schema": {
      "TOOL_REQUIRED": "Tool that was planned/required (if explicitly stated)",
      "FABRICATED_CLAIMS": "Specific claims that are unsupported by any tool observation",
      "ALLOWED_EVIDENCE": "Observations present in the trace context (if available)",
      "SAFE_REWRITE_MODE": "Either 'remove claim' or 'mark unknown and request tool call'"
    },
    "error_type_spec_text": "error_type: Tool-related\ntrail_definition: Fabricated the outcome of a tool interaction or data retrieval step / fabricated tool capabilities.\npatch_side_default: replace_span_output\nslot_extraction:\n- TOOL_REQUIRED: extract any required tool name from plan/instructions if present.\n- FABRICATED_CLAIMS: extract the unsupported assertions from ERROR_EVIDENCE.\n- ALLOWED_EVIDENCE: extract any actual observations present in local context.\n- SAFE_REWRITE_MODE: default to removing fabricated claims; if required, request a tool call instead of asserting.\nrepair_instruction:\n- Rewrite the span to remove FABRICATED_CLAIMS.\n- If the task requires those facts, replace assertions with a request to call TOOL_REQUIRED (or an explicit 'need to look up').\n- Keep minimal edits; do not introduce new facts.\nforbidden_actions:\n- Do not invent tool observations.\n- Do not invent citations or sources.\n- Do not convert the span into a full new plan unless it is already a planning span.\npostcheck:\n- FABRICATED_CLAIMS are removed or rephrased as unknown/needs tool.\n- No new factual claims beyond ALLOWED_EVIDENCE.\n- Patch does not add ungrounded tokens.",
    "repair_instruction": "Rewrite the span to remove fabricated tool-derived claims. If the information is required, replace assertions with an explicit need to call the required tool instead of claiming results.",
    "forbidden_actions": [
      "Do not invent tool observations.",
      "Do not invent citations or sources.",
      "Do not convert the span into a full new plan unless it is already a planning span."
    ],
    "postcheck": [
      "Fabricated claims are removed or rewritten as unknown/needs tool.",
      "No new factual claims beyond allowed evidence.",
      "No ungrounded special tokens introduced.",
      "patch_payload differs from local_snippet."
    ]
  },
  "Language-only": {
    "category": "Language-only",
    "trail_definition": "Ungrounded language-only hallucination (not tied to tool use).",
    "patch_side_default": "replace_span_output",
    "slot_schema": {
      "UNSUPPORTED_CLAIMS": "Claims not supported by provided context/task",
      "CONTEXT_GIVEN": "Facts explicitly present in the prompt/context",
      "SAFE_REWRITE_MODE": "Either 'remove claim' or 'hedge + request verification'"
    },
    "error_type_spec_text": "error_type: Language-only\ntrail_definition: Ungrounded language-only hallucination (not tied to tool use).\npatch_side_default: replace_span_output\nslot_extraction:\n- UNSUPPORTED_CLAIMS: extract claims identified in ERROR_EVIDENCE.\n- CONTEXT_GIVEN: extract facts explicitly present in USER_REQUIREMENTS/local context.\n- SAFE_REWRITE_MODE: default to removing unsupported claims; optionally hedge + request verification.\nrepair_instruction:\n- Rewrite the span to remove or hedge UNSUPPORTED_CLAIMS.\n- Keep only statements grounded in CONTEXT_GIVEN.\n- If the task cannot be completed without the missing fact, explicitly state it needs lookup (without inventing).\nforbidden_actions:\n- Do not add new facts.\n- Do not add invented sources.\n- Do not rewrite unrelated parts of the output.\npostcheck:\n- Unsupported claims removed/hedged.\n- Remaining content is grounded in CONTEXT_GIVEN.\n- No new unsupported factual content.",
    "repair_instruction": "Rewrite the span to remove or hedge unsupported claims and keep only statements grounded in provided context. If missing info is required, explicitly state it needs lookup.",
    "forbidden_actions": [
      "Do not add new facts.",
      "Do not add invented sources.",
      "Do not rewrite unrelated parts of the output."
    ],
    "postcheck": [
      "Unsupported claims removed or hedged.",
      "Remaining content is grounded in provided context.",
      "No new unsupported factual content introduced.",
      "patch_payload differs from local_snippet."
    ]
  },
  "Goal Deviation": {
    "category": "Goal Deviation",
    "trail_definition": "Deviated from the task or subtask (did something else).",
    "patch_side_default": "replace_span_input",
    "slot_schema": {
      "ACTIVE_GOAL": "Current task/subtask objective explicitly stated",
      "DEVIATION_BEHAVIOR": "What the span is doing instead (from ERROR_EVIDENCE)",
      "NEXT_REQUIRED_ACTION": "Minimal next step aligned with ACTIVE_GOAL",
      "STOP_IF_UNCERTAIN": "Rule to avoid continuing off-goal"
    },
    "error_type_spec_text": "error_type: Goal Deviation\ntrail_definition: Deviated from the task or subtask.\npatch_side_default: replace_span_input\nslot_extraction:\n- ACTIVE_GOAL: extract the explicit task/subtask objective.\n- DEVIATION_BEHAVIOR: extract what it did instead.\n- NEXT_REQUIRED_ACTION: infer the minimal next aligned action (tool call, check, or step).\n- STOP_IF_UNCERTAIN: add a conservative redirect rule.\nrepair_instruction:\n- Rewrite the local instruction/thought to restate ACTIVE_GOAL and require NEXT_REQUIRED_ACTION.\n- Add a single-line redirect: if the next action does not advance ACTIVE_GOAL, do not proceed.\n- Keep the patch local.\nforbidden_actions:\n- Do not fabricate tool outputs or results.\n- Do not add new subtasks.\n- Do not rewrite the entire plan; only correct the immediate deviation.\npostcheck:\n- ACTIVE_GOAL is explicitly stated.\n- NEXT_REQUIRED_ACTION is aligned with ACTIVE_GOAL.\n- No new unsupported facts were added.",
    "repair_instruction": "Rewrite the local instruction/thought to restate the active goal and require the minimal next aligned action. Add a conservative redirect rule to avoid off-goal actions.",
    "forbidden_actions": [
      "Do not fabricate tool outputs or results.",
      "Do not add new subtasks.",
      "Do not rewrite the entire plan; only correct the immediate deviation."
    ],
    "postcheck": [
      "Active goal is explicitly stated.",
      "Next required action aligns with active goal.",
      "No new unsupported facts were added.",
      "patch_payload differs from local_snippet."
    ]
  },
  "Context Handling Failures": {
    "category": "Context Handling Failures",
    "trail_definition": "Window overflow / state tracking / forgetting important context.",
    "patch_side_default": "replace_span_input",
    "slot_schema": {
      "CRITICAL_FACTS": "Facts/constraints that must persist (from USER_REQUIREMENTS or earlier context)",
      "CURRENT_STATE": "Variables/choices already made (if explicit)",
      "MISSING_CONTEXT": "What was forgotten (from ERROR_EVIDENCE)",
      "PIN_RULE": "Instruction to keep these facts in working memory"
    },
    "error_type_spec_text": "error_type: Context Handling Failures\ntrail_definition: Window overflow / state tracking / forgetting important context.\npatch_side_default: replace_span_input\nslot_extraction:\n- CRITICAL_FACTS: extract must-remember constraints and previously established facts.\n- CURRENT_STATE: extract explicit variables/decisions.\n- MISSING_CONTEXT: extract the forgotten item from ERROR_EVIDENCE.\n- PIN_RULE: add a short 'pin' instruction.\nrepair_instruction:\n- Rewrite the local input context to include a concise pinned summary of CRITICAL_FACTS and CURRENT_STATE.\n- Add a rule: do not proceed if MISSING_CONTEXT is required but not present; request it.\n- Keep it minimal: only include facts needed for the next step.\nforbidden_actions:\n- Do not invent missing facts.\n- Do not add new requirements.\n- Do not summarize the entire conversation; pin only critical items.\npostcheck:\n- Pinned summary includes grounded CRITICAL_FACTS.\n- No fabricated content.\n- Patch is minimal and local.",
    "repair_instruction": "Inject a minimal pinned summary of critical facts/state into the local input and add a rule to stop/request clarification if required context is missing.",
    "forbidden_actions": [
      "Do not invent missing facts.",
      "Do not add new requirements.",
      "Do not summarize the entire conversation; pin only critical items."
    ],
    "postcheck": [
      "Pinned summary includes grounded critical facts/state.",
      "No fabricated content introduced.",
      "Patch is minimal and local.",
      "patch_payload differs from local_snippet."
    ]
  },
  "Tool Definition Issues": {
    "category": "Tool Definition Issues",
    "trail_definition": "Tool defined incorrectly / inconsistent with its description or required schema.",
    "patch_side_default": "replace_span_input",
    "slot_schema": {
      "TOOL_NAME": "Tool name in local context",
      "TOOL_SPEC_TEXT": "The tool description/parameters text in local context",
      "SCHEMA_MISMATCH": "What is wrong (missing arg, wrong arg name/type) from ERROR_EVIDENCE",
      "CORRECTED_CALL_PATTERN": "Correct function signature usage grounded in TOOL_SPEC_TEXT"
    },
    "error_type_spec_text": "error_type: Tool Definition Issues\ntrail_definition: Tool defined incorrectly / inconsistent with description.\npatch_side_default: replace_span_input\nslot_extraction:\n- TOOL_NAME: extract tool name.\n- TOOL_SPEC_TEXT: extract the tool parameters schema from context.\n- SCHEMA_MISMATCH: extract mismatch from ERROR_EVIDENCE.\n- CORRECTED_CALL_PATTERN: derive correct call pattern using only TOOL_SPEC_TEXT.\nrepair_instruction:\n- Rewrite the local span to align tool invocation with TOOL_SPEC_TEXT (correct argument names/types).\n- If LOCAL_SNIPPET is tool_calls JSON, fix only function.name or function.arguments.\n- CRITICAL: Only valid keys per tool_call entry are: id, type, function. Do not add phantom keys.\nforbidden_actions:\n- Do not add phantom keys to tool_calls entries.\n- Do not add new tools not in context.\n- Do not fabricate tool outputs.\npostcheck:\n- Tool call matches TOOL_SPEC_TEXT.\n- tool_calls JSON schema preserved (only id/type/function).\n- No fabricated outputs.",
    "repair_instruction": "Align tool invocation with the grounded tool specification by correcting only the tool call name/arguments (or the local instruction that produces them).",
    "forbidden_actions": [
      "Do not add phantom keys to tool_calls entries.",
      "Do not add new tools not in context.",
      "Do not fabricate tool outputs."
    ],
    "postcheck": [
      "Tool call matches grounded tool spec (arg names/types).",
      "tool_calls JSON schema preserved (only id/type/function keys).",
      "No fabricated outputs introduced.",
      "patch_payload differs from local_snippet."
    ]
  },
  "Environment Setup Errors": {
    "category": "Environment Setup Errors",
    "trail_definition": "Permission/resource/API-key setup failures; missing environment prerequisites.",
    "patch_side_default": "replace_span_input",
    "slot_schema": {
      "MISSING_PREREQ": "What prerequisite is missing (from ERROR_EVIDENCE)",
      "ACCESS_CONSTRAINT": "Permission or setup constraint if explicit",
      "SAFE_FAIL_ACTION": "Abort with explicit error, or request user-provided credential, or switch to alternative approach"
    },
    "error_type_spec_text": "error_type: Environment Setup Errors\ntrail_definition: Permission/resource/API-key setup failures.\npatch_side_default: replace_span_input\nslot_extraction:\n- MISSING_PREREQ: extract missing prerequisite from ERROR_EVIDENCE.\n- ACCESS_CONSTRAINT: extract any explicit permission/setup constraint.\n- SAFE_FAIL_ACTION: choose conservative fallback (request setup, or stop).\nrepair_instruction:\n- Rewrite local span to perform a pre-check for MISSING_PREREQ and avoid proceeding without it.\n- Replace failing action with SAFE_FAIL_ACTION (e.g., ask for credential, switch to non-auth method if allowed).\nforbidden_actions:\n- Do not invent credentials.\n- Do not claim setup succeeded.\n- Do not add new tools.\npostcheck:\n- Includes explicit recognition of missing prerequisite and a safe fallback.\n- No invented credentials or success claims.",
    "repair_instruction": "Add a local pre-check for missing prerequisites and replace the failing action with a safe fallback (request setup/credentials or stop), without inventing credentials.",
    "forbidden_actions": [
      "Do not invent credentials.",
      "Do not claim setup succeeded.",
      "Do not add new tools."
    ],
    "postcheck": [
      "Explicitly recognizes missing prerequisite and uses safe fallback.",
      "No invented credentials or success claims.",
      "patch_payload differs from local_snippet."
    ]
  },
  "Rate Limiting": {
    "category": "Rate Limiting",
    "trail_definition": "Rate limiting errors (e.g., 429) preventing tool/API completion.",
    "patch_side_default": "replace_span_input",
    "slot_schema": {
      "TOOL_NAME": "Tool/API being called when rate limited",
      "RETRY_POLICY": "Backoff policy grounded in error evidence (e.g., wait and retry once)",
      "MAX_RETRIES": "Conservative retry count (default 1 unless grounded)",
      "FALLBACK_ACTION": "Alternative plan if retry fails (summarize what is missing; ask user; stop)"
    },
    "error_type_spec_text": "error_type: Rate Limiting\ntrail_definition: Rate limiting errors (e.g., 429).\npatch_side_default: replace_span_input\nslot_extraction:\n- TOOL_NAME: extract tool being called.\n- RETRY_POLICY: set conservative backoff (sleep/wait) if allowed; otherwise 'retry once later'.\n- MAX_RETRIES: default 1 unless explicitly grounded.\n- FALLBACK_ACTION: choose a safe fallback.\nrepair_instruction:\n- Rewrite the local step to avoid immediate repeated identical calls.\n- Add a single retry with backoff and a clear fallback if still rate limited.\n- Keep changes minimal and local.\n- CRITICAL: Only valid keys per tool_call entry are: id, type, function. Do not add retry_limit or backoff as tool_call fields.\nforbidden_actions:\n- Do not invent successful tool results.\n- Do not add schema-invalid fields to tool_calls.\n- Do not implement complex loops.\npostcheck:\n- Removes immediate repeated identical call.\n- Includes bounded retry and fallback.\n- No fabricated outputs.",
    "repair_instruction": "Avoid immediate repeated identical calls; add one bounded retry with backoff and a clear fallback if rate limiting persists, without fabricating results.",
    "forbidden_actions": [
      "Do not invent successful tool results.",
      "Do not add schema-invalid fields to tool_calls.",
      "Do not implement complex loops."
    ],
    "postcheck": [
      "Prevents immediate repeated identical call.",
      "Includes bounded retry and fallback.",
      "No fabricated outputs introduced.",
      "patch_payload differs from local_snippet."
    ]
  },
  "Authentication Errors": {
    "category": "Authentication Errors",
    "trail_definition": "Authentication/permission errors (e.g., 401/403).",
    "patch_side_default": "replace_span_input",
    "slot_schema": {
      "TOOL_NAME": "Tool/API requiring auth",
      "AUTH_REQUIREMENT": "What credential/scope is required if explicitly stated",
      "FAILSAFE_ACTION": "Ask for credentials, switch to public source, or stop"
    },
    "error_type_spec_text": "error_type: Authentication Errors\ntrail_definition: Authentication/permission errors (401/403).\npatch_side_default: replace_span_input\nslot_extraction:\n- TOOL_NAME: extract tool/API name.\n- AUTH_REQUIREMENT: extract any explicit credential/scope requirement.\n- FAILSAFE_ACTION: select conservative fallback.\nrepair_instruction:\n- Rewrite local span to stop attempting unauthorized calls.\n- Replace with FAILSAFE_ACTION: request credential or switch to an allowed public alternative if explicitly permitted.\n- CRITICAL: Only valid keys per tool_call entry are: id, type, function. Do not add auth fields.\nforbidden_actions:\n- Do not invent credentials.\n- Do not claim authentication succeeded.\n- Do not fabricate tool output.\npostcheck:\n- No unauthorized call is retried without new credentials.\n- Contains a safe fallback.\n- No fabricated outputs.",
    "repair_instruction": "Stop retrying unauthorized calls; replace with a safe fallback (request credentials or switch to allowed public alternative), without inventing authentication success.",
    "forbidden_actions": [
      "Do not invent credentials.",
      "Do not claim authentication succeeded.",
      "Do not fabricate tool output."
    ],
    "postcheck": [
      "Does not retry unauthorized call without new credentials.",
      "Contains safe fallback behavior.",
      "No fabricated outputs introduced.",
      "patch_payload differs from local_snippet."
    ]
  },
  "Service Errors": {
    "category": "Service Errors",
    "trail_definition": "Service-side errors (e.g., 5xx) from tools/APIs.",
    "patch_side_default": "replace_span_input",
    "slot_schema": {
      "TOOL_NAME": "Tool/API experiencing 5xx",
      "RETRY_ONCE": "Whether to retry once (default yes)",
      "FALLBACK_ACTION": "Switch tool/source or stop with explicit missing info"
    },
    "error_type_spec_text": "error_type: Service Errors\ntrail_definition: Service-side errors (5xx).\npatch_side_default: replace_span_input\nslot_extraction:\n- TOOL_NAME: extract tool name.\n- RETRY_ONCE: default true.\n- FALLBACK_ACTION: choose conservative fallback.\nrepair_instruction:\n- Add at most one retry (non-identical if possible) and then a fallback.\n- Do not loop.\n- CRITICAL: Only valid keys per tool_call entry are: id, type, function. Do not add retry or fallback as tool_call fields.\nforbidden_actions:\n- Do not fabricate tool results.\n- Do not add invalid tool_calls keys.\n- Do not mask the error.\npostcheck:\n- Includes bounded retry then fallback.\n- No fabricated outputs.",
    "repair_instruction": "Add at most one retry for transient service errors, then switch to a fallback action; do not fabricate results or loop.",
    "forbidden_actions": [
      "Do not fabricate tool results.",
      "Do not add invalid tool_calls keys.",
      "Do not mask the error."
    ],
    "postcheck": [
      "Bounded retry then fallback is present.",
      "No fabricated outputs introduced.",
      "patch_payload differs from local_snippet."
    ]
  },
  "Resource Not Found": {
    "category": "Resource Not Found",
    "trail_definition": "Missing resource/path/id (e.g., 404) or nonexistent file/resource reference.",
    "patch_side_default": "replace_span_input",
    "slot_schema": {
      "RESOURCE_ID": "The referenced id/path/url",
      "LOOKUP_STRATEGY": "How to find correct resource (search, list, inspect directory) if grounded",
      "FALLBACK_ACTION": "Ask user for correct id/path or stop"
    },
    "error_type_spec_text": "error_type: Resource Not Found\ntrail_definition: Missing resource/path/id (404).\npatch_side_default: replace_span_input\nslot_extraction:\n- RESOURCE_ID: extract referenced id/path/url.\n- LOOKUP_STRATEGY: derive from available tools/context.\n- FALLBACK_ACTION: request correct identifier if necessary.\nrepair_instruction:\n- Rewrite local step to verify RESOURCE_ID exists before using it.\n- If not verifiable, invoke LOOKUP_STRATEGY or FALLBACK_ACTION.\n- CRITICAL: Only valid keys per tool_call entry are: id, type, function.\nforbidden_actions:\n- Do not invent a new resource id.\n- Do not claim a resource exists without evidence.\n- Do not fabricate retrieval results.\npostcheck:\n- Includes explicit existence check or lookup.\n- No invented identifiers.",
    "repair_instruction": "Verify resource existence before use; if not verifiable, perform a grounded lookup or request the correct identifier instead of guessing.",
    "forbidden_actions": [
      "Do not invent a new resource id.",
      "Do not claim a resource exists without evidence.",
      "Do not fabricate retrieval results."
    ],
    "postcheck": [
      "Includes existence check or grounded lookup step.",
      "No invented identifiers or fabricated results.",
      "patch_payload differs from local_snippet."
    ]
  },
  "Resource Exhaustion": {
    "category": "Resource Exhaustion",
    "trail_definition": "Resource exhaustion (memory/compute) including memory overflow.",
    "patch_side_default": "replace_span_input",
    "slot_schema": {
      "BUDGET_TYPE": "Memory/token/compute budget that was exceeded",
      "CHUNKING_RULE": "Chunk size or batching rule",
      "MINIMAL_SUBGOAL": "Next minimal computation/retrieval step",
      "STOP_CONDITION": "When to stop and summarize partial results"
    },
    "error_type_spec_text": "error_type: Resource Exhaustion\ntrail_definition: Resource exhaustion (memory/compute), including memory overflow.\npatch_side_default: replace_span_input\nslot_extraction:\n- BUDGET_TYPE: extract which budget failed if stated.\n- CHUNKING_RULE: set conservative chunking/batching.\n- MINIMAL_SUBGOAL: infer minimal next step.\n- STOP_CONDITION: define when to stop.\nrepair_instruction:\n- Rewrite local step to reduce resource use via chunking, limiting scope, or summarizing intermediate results.\n- Add a bounded plan: process only a small unit at a time.\nforbidden_actions:\n- Do not expand scope.\n- Do not add repeated tool loops.\n- Do not fabricate skipped outputs.\npostcheck:\n- Includes explicit chunking/limit.\n- Scope is reduced, not expanded.\n- No fabricated results.",
    "repair_instruction": "Reduce resource usage by chunking/limiting scope and adding a bounded processing plan; do not expand scope or fabricate results.",
    "forbidden_actions": [
      "Do not expand scope.",
      "Do not add repeated tool loops.",
      "Do not fabricate skipped outputs."
    ],
    "postcheck": [
      "Includes explicit chunking/limit mechanism.",
      "Scope reduced rather than expanded.",
      "No fabricated results introduced.",
      "patch_payload differs from local_snippet."
    ]
  },
  "Timeout Issues": {
    "category": "Timeout Issues",
    "trail_definition": "System took too long to respond / tool or process timed out.",
    "patch_side_default": "replace_span_input",
    "slot_schema": {
      "TIME_BUDGET": "Any explicit timeout threshold if present",
      "SIMPLIFICATION": "How to simplify the action (smaller query, fewer pages, smaller file range)",
      "MAX_ATTEMPTS": "Bounded retry count (default 1)",
      "FALLBACK_ACTION": "Stop and summarize what is missing / alternative source"
    },
    "error_type_spec_text": "error_type: Timeout Issues\ntrail_definition: Tool/process timed out.\npatch_side_default: replace_span_input\nslot_extraction:\n- TIME_BUDGET: extract any explicit time budget.\n- SIMPLIFICATION: choose a simpler/smaller action.\n- MAX_ATTEMPTS: default 1.\n- FALLBACK_ACTION: choose a safe fallback.\nrepair_instruction:\n- Rewrite the local step to reduce workload and avoid long operations.\n- Add bounded retry and fallback.\n- CRITICAL: Only valid keys per tool_call entry are: id, type, function.\nforbidden_actions:\n- Do not loop.\n- Do not fabricate results.\n- Do not increase scope.\npostcheck:\n- Action is simplified.\n- Retry is bounded.\n- No fabricated results.",
    "repair_instruction": "Simplify the local action to avoid long operations, add a bounded retry and fallback, and do not fabricate results.",
    "forbidden_actions": [
      "Do not loop.",
      "Do not fabricate results.",
      "Do not increase scope."
    ],
    "postcheck": [
      "Action simplified relative to original.",
      "Retry bounded and fallback present.",
      "No fabricated results introduced.",
      "patch_payload differs from local_snippet."
    ]
  }
}
\end{codebox}

\begin{codebox}{MAST patch library (\texttt{causal\_graph/causal\_valid/patch\_library.json})}
{
  "1.1": {
    "category": "1.1",
    "mast_definition": "Disobey Task Specification \u2014 Violates constraints or requirements explicitly stated in the task.",
    "patch_side_default": "replace_step_content",
    "slot_schema": {
      "VIOLATED_CONSTRAINT": "The specific constraint or requirement that was violated (quoted from task or evidence)",
      "CORRECT_BEHAVIOR": "What the agent should have done to comply with the constraint"
    },
    "error_type_spec_text": "error_type: 1.1 Disobey Task Specification\nmast_definition: Violates constraints or requirements explicitly stated in the task.\npatch_side_default: replace_step_content\nslot_extraction:\n- VIOLATED_CONSTRAINT: identify the specific requirement or constraint from the task that was violated, quoting it verbatim from ERROR_EVIDENCE or USER_REQUIREMENTS.\n- CORRECT_BEHAVIOR: describe the correct behavior that satisfies the violated constraint.\nrepair_instruction:\n- Rewrite the step content so that it complies with the stated constraint.\n- Preserve all other content and reasoning that does not violate any constraint.\n- Do not introduce new capabilities or expand the solution scope.\nforbidden_actions:\n- Do not remove valid reasoning or correct steps.\n- Do not fabricate new task requirements.\n- Do not change the solution approach unless required to satisfy the constraint.\npostcheck:\n- The violated constraint is no longer present in the patched step.\n- patch_payload differs from local_snippet.",
    "repair_instruction": "Rewrite the step so that it complies with the stated task constraint. Keep all valid reasoning intact.",
    "forbidden_actions": [
      "Do not fabricate new task requirements.",
      "Do not change the solution approach unless required to satisfy the constraint."
    ],
    "postcheck": [
      "The violated constraint is no longer present.",
      "patch_payload differs from local_snippet."
    ]
  },
  "1.2": {
    "category": "1.2",
    "mast_definition": "Disobey Role Specification \u2014 Ignores or violates the assigned agent role.",
    "patch_side_default": "replace_step_content",
    "slot_schema": {
      "ROLE_REQUIREMENT": "The role specification that was violated",
      "CORRECT_ROLE_BEHAVIOR": "What behavior is consistent with the assigned role"
    },
    "error_type_spec_text": "error_type: 1.2 Disobey Role Specification\nmast_definition: Ignores or violates the assigned agent role.\npatch_side_default: replace_step_content\nslot_extraction:\n- ROLE_REQUIREMENT: identify the role specification being violated from ERROR_DESCRIPTION or USER_REQUIREMENTS.\n- CORRECT_ROLE_BEHAVIOR: describe the behavior that is consistent with the assigned role.\nrepair_instruction:\n- Rewrite the step so that the agent behaves within its assigned role.\n- Preserve the task-relevant content; only adjust the role-violating aspects.\nforbidden_actions:\n- Do not change the agent's goal or task objective.\n- Do not fabricate role specifications.\npostcheck:\n- The role violation is resolved in the patched step.\n- patch_payload differs from local_snippet.",
    "repair_instruction": "Rewrite the step so the agent operates within its assigned role.",
    "forbidden_actions": [
      "Do not change the agent's goal or task objective.",
      "Do not fabricate role specifications."
    ],
    "postcheck": [
      "The role violation is resolved.",
      "patch_payload differs from local_snippet."
    ]
  },
  "1.3": {
    "category": "1.3",
    "mast_definition": "Step Repetition \u2014 Repeats a task or phase already completed with a result.",
    "patch_side_default": "replace_step_content",
    "slot_schema": {
      "REPEATED_STEP": "Description of the step or phase that was unnecessarily repeated",
      "PRIOR_RESULT": "The result that was already achieved before the repetition"
    },
    "error_type_spec_text": "error_type: 1.3 Step Repetition\nmast_definition: Repeats a task or phase already completed with a result.\npatch_side_default: replace_step_content\nslot_extraction:\n- REPEATED_STEP: identify what step or phase was unnecessarily repeated from ERROR_DESCRIPTION.\n- PRIOR_RESULT: identify the result that was already obtained before this repetition.\nrepair_instruction:\n- Replace the repetition with an acknowledgment of the prior result and a forward-looking step.\n- The agent should reference the previous result and proceed to the next logical step instead of repeating.\nforbidden_actions:\n- Do not fabricate new prior results.\n- Do not skip required steps (only remove genuinely redundant repetition).\npostcheck:\n- The repeated content is removed or replaced.\n- patch_payload differs from local_snippet.",
    "repair_instruction": "Replace the repeated step with acknowledgment of the prior result and a forward move.",
    "forbidden_actions": [
      "Do not fabricate prior results.",
      "Do not skip genuinely required steps."
    ],
    "postcheck": [
      "The redundant repetition is removed.",
      "patch_payload differs from local_snippet."
    ]
  },
  "1.4": {
    "category": "1.4",
    "mast_definition": "Loss of Conversation History \u2014 Fails to retain or use prior conversation context.",
    "patch_side_default": "replace_step_content",
    "slot_schema": {
      "LOST_CONTEXT": "The specific prior information that was not retained or referenced",
      "REQUIRED_REFERENCE": "How the prior context should have been used in this step"
    },
    "error_type_spec_text": "error_type: 1.4 Loss of Conversation History\nmast_definition: Fails to retain or use prior conversation context.\npatch_side_default: replace_step_content\nslot_extraction:\n- LOST_CONTEXT: identify the prior information that was ignored or forgotten from ERROR_EVIDENCE.\n- REQUIRED_REFERENCE: describe how the prior context should have been incorporated.\nrepair_instruction:\n- Rewrite the step to correctly reference and incorporate the relevant prior context.\n- The agent should explicitly acknowledge the prior information and build on it.\nforbidden_actions:\n- Do not fabricate prior conversation content.\n- Do not introduce information that was not in the prior conversation.\npostcheck:\n- The step correctly references or incorporates the prior context.\n- patch_payload differs from local_snippet.",
    "repair_instruction": "Rewrite the step so it correctly references and uses the relevant prior context.",
    "forbidden_actions": [
      "Do not fabricate prior conversation content.",
      "Do not introduce information not in prior context."
    ],
    "postcheck": [
      "Prior context is correctly referenced.",
      "patch_payload differs from local_snippet."
    ]
  },
  "1.5": {
    "category": "1.5",
    "mast_definition": "Unaware of Termination Conditions \u2014 Continues past a valid stopping point or stops too early.",
    "patch_side_default": "replace_step_content",
    "slot_schema": {
      "TERMINATION_CONDITION": "The stopping condition that was missed or misidentified",
      "CORRECT_ACTION": "Whether the agent should stop or continue, and why"
    },
    "error_type_spec_text": "error_type: 1.5 Unaware of Termination Conditions\nmast_definition: Continues past a valid stopping point or stops too early.\npatch_side_default: replace_step_content\nslot_extraction:\n- TERMINATION_CONDITION: identify the termination condition from ERROR_DESCRIPTION.\n- CORRECT_ACTION: determine whether the agent should stop (if it continued too long) or continue (if it stopped too early).\nrepair_instruction:\n- If the agent continued past a valid stop: rewrite to acknowledge completion and terminate cleanly.\n- If the agent stopped too early: rewrite to continue toward the required completion criterion.\nforbidden_actions:\n- Do not fabricate completion criteria.\n- Do not alter the task goal.\npostcheck:\n- The termination behavior is corrected.\n- patch_payload differs from local_snippet.",
    "repair_instruction": "Correct the termination behavior: stop if past valid stopping point, continue if stopped too early.",
    "forbidden_actions": [
      "Do not fabricate completion criteria.",
      "Do not alter the task goal."
    ],
    "postcheck": [
      "Termination behavior is corrected.",
      "patch_payload differs from local_snippet."
    ]
  },
  "2.1": {
    "category": "2.1",
    "mast_definition": "Conversation Reset \u2014 Resets the conversation, losing prior context and progress.",
    "patch_side_default": "replace_step_content",
    "slot_schema": {
      "LOST_PROGRESS": "The prior context or progress that was discarded by the reset",
      "RECOVERY_ACTION": "How the step should be rewritten to preserve prior context"
    },
    "error_type_spec_text": "error_type: 2.1 Conversation Reset\nmast_definition: Resets the conversation, losing prior context and progress.\npatch_side_default: replace_step_content\nslot_extraction:\n- LOST_PROGRESS: identify the prior context or progress that the reset discarded.\n- RECOVERY_ACTION: describe how to rewrite the step to preserve context continuity.\nrepair_instruction:\n- Rewrite the step to continue from the prior state instead of resetting.\n- Explicitly reference the prior context that would have been lost.\nforbidden_actions:\n- Do not fabricate prior progress.\n- Do not remove legitimately new information.\npostcheck:\n- The reset is avoided; prior context is referenced.\n- patch_payload differs from local_snippet.",
    "repair_instruction": "Rewrite to continue from the prior state rather than resetting.",
    "forbidden_actions": [
      "Do not fabricate prior progress.",
      "Do not remove legitimately new information."
    ],
    "postcheck": [
      "The conversation reset is avoided.",
      "patch_payload differs from local_snippet."
    ]
  },
  "2.2": {
    "category": "2.2",
    "mast_definition": "Fail to Ask for Clarification \u2014 Proceeds without resolving ambiguity that required clarification.",
    "patch_side_default": "replace_step_content",
    "slot_schema": {
      "AMBIGUITY": "The specific ambiguity or unclear aspect that required clarification",
      "CLARIFICATION_QUESTION": "A concrete clarification question the agent should have asked"
    },
    "error_type_spec_text": "error_type: 2.2 Fail to Ask for Clarification\nmast_definition: Proceeds without resolving ambiguity that required clarification.\npatch_side_default: replace_step_content\nslot_extraction:\n- AMBIGUITY: identify the specific ambiguity from ERROR_DESCRIPTION or ERROR_EVIDENCE.\n- CLARIFICATION_QUESTION: formulate the question the agent should have asked before proceeding.\nrepair_instruction:\n- Rewrite the step to explicitly identify the ambiguity and ask for clarification before proceeding.\n- The agent should pause its current action and request the missing information.\nforbidden_actions:\n- Do not fabricate an answer to the ambiguity.\n- Do not proceed with the original action while just adding a clarification note.\npostcheck:\n- The step asks for clarification rather than proceeding with the ambiguous assumption.\n- patch_payload differs from local_snippet.",
    "repair_instruction": "Rewrite to ask for clarification instead of proceeding with an ambiguous assumption.",
    "forbidden_actions": [
      "Do not fabricate an answer to the ambiguity.",
      "Do not proceed with the original action while just noting the ambiguity."
    ],
    "postcheck": [
      "The step asks for clarification before proceeding.",
      "patch_payload differs from local_snippet."
    ]
  },
  "2.3": {
    "category": "2.3",
    "mast_definition": "Task Derailment \u2014 Shifts focus away from the intended objective.",
    "patch_side_default": "replace_step_content",
    "slot_schema": {
      "INTENDED_OBJECTIVE": "The original task objective that was abandoned or deprioritized",
      "DERAILMENT_DESCRIPTION": "How the agent shifted focus away from the objective"
    },
    "error_type_spec_text": "error_type: 2.3 Task Derailment\nmast_definition: Shifts focus away from the intended objective.\npatch_side_default: replace_step_content\nslot_extraction:\n- INTENDED_OBJECTIVE: identify the main task objective from USER_REQUIREMENTS or ERROR_DESCRIPTION.\n- DERAILMENT_DESCRIPTION: describe how the agent drifted from that objective.\nrepair_instruction:\n- Rewrite the step to refocus on the intended objective.\n- Explicitly reference the main goal and orient the step toward it.\nforbidden_actions:\n- Do not introduce new objectives.\n- Do not remove relevant subtasks that genuinely serve the main objective.\npostcheck:\n- The step is focused on the intended objective.\n- patch_payload differs from local_snippet.",
    "repair_instruction": "Rewrite to refocus on the intended objective.",
    "forbidden_actions": [
      "Do not introduce new objectives.",
      "Do not remove subtasks that serve the main goal."
    ],
    "postcheck": [
      "The step is refocused on the intended objective.",
      "patch_payload differs from local_snippet."
    ]
  },
  "2.4": {
    "category": "2.4",
    "mast_definition": "Information Withholding \u2014 Fails to share information needed by other agents.",
    "patch_side_default": "replace_step_content",
    "slot_schema": {
      "WITHHELD_INFORMATION": "The specific information that was not shared",
      "WHY_NEEDED": "Why the other agent needs this information"
    },
    "error_type_spec_text": "error_type: 2.4 Information Withholding\nmast_definition: Fails to share information needed by other agents.\npatch_side_default: replace_step_content\nslot_extraction:\n- WITHHELD_INFORMATION: identify the information that was not passed to the other agent.\n- WHY_NEEDED: explain why the receiving agent needs this information to complete the task.\nrepair_instruction:\n- Rewrite the step to include and explicitly communicate the withheld information.\n- The information must be presented clearly for the receiving agent to use.\nforbidden_actions:\n- Do not fabricate information that was not available.\n- Do not add unrelated information.\npostcheck:\n- The required information is now shared in the patched step.\n- patch_payload differs from local_snippet.",
    "repair_instruction": "Rewrite to include the withheld information for the receiving agent.",
    "forbidden_actions": [
      "Do not fabricate information that was not available.",
      "Do not add unrelated information."
    ],
    "postcheck": [
      "The required information is now communicated.",
      "patch_payload differs from local_snippet."
    ]
  },
  "2.6": {
    "category": "2.6",
    "mast_definition": "Action-Reasoning Mismatch \u2014 Executes an action inconsistent with the stated reasoning.",
    "patch_side_default": "replace_step_content",
    "slot_schema": {
      "STATED_REASONING": "The reasoning or plan that was stated",
      "INCONSISTENT_ACTION": "The action that contradicts the stated reasoning"
    },
    "error_type_spec_text": "error_type: 2.6 Action-Reasoning Mismatch\nmast_definition: Executes an action inconsistent with the stated reasoning.\npatch_side_default: replace_step_content\nslot_extraction:\n- STATED_REASONING: identify the reasoning or plan from ERROR_EVIDENCE or LOCAL_SNIPPET.\n- INCONSISTENT_ACTION: identify the action that contradicts the stated reasoning.\nrepair_instruction:\n- Rewrite the step so the action is consistent with the stated reasoning.\n- Either update the reasoning to match the action, or update the action to match the reasoning \u2014 whichever requires less change.\nforbidden_actions:\n- Do not fabricate new reasoning.\n- Do not change the overall task goal.\npostcheck:\n- The action and reasoning are now consistent in the patched step.\n- patch_payload differs from local_snippet.",
    "repair_instruction": "Align the action with the stated reasoning (or vice versa) with minimal change.",
    "forbidden_actions": [
      "Do not fabricate new reasoning.",
      "Do not change the overall task goal."
    ],
    "postcheck": [
      "Action and reasoning are now consistent.",
      "patch_payload differs from local_snippet."
    ]
  },
  "3.1": {
    "category": "3.1",
    "mast_definition": "Premature Termination \u2014 Stops before the task is complete or a required output is produced.",
    "patch_side_default": "replace_step_content",
    "slot_schema": {
      "INCOMPLETE_ASPECT": "What part of the task was left incomplete",
      "REQUIRED_CONTINUATION": "What the agent should have continued doing"
    },
    "error_type_spec_text": "error_type: 3.1 Premature Termination\nmast_definition: Stops before the task is complete or a required output is produced.\npatch_side_default: replace_step_content\nslot_extraction:\n- INCOMPLETE_ASPECT: identify what was left unfinished from ERROR_DESCRIPTION.\n- REQUIRED_CONTINUATION: describe what the agent should have continued doing instead of stopping.\nrepair_instruction:\n- Rewrite the step to continue the task rather than terminate.\n- Explicitly identify remaining work and commit to completing it.\nforbidden_actions:\n- Do not fabricate completion of the remaining work in this step.\n- Do not add work beyond what was originally required.\npostcheck:\n- The step continues toward task completion rather than terminating.\n- patch_payload differs from local_snippet.",
    "repair_instruction": "Rewrite to continue the task rather than terminate prematurely.",
    "forbidden_actions": [
      "Do not fabricate completion of remaining work.",
      "Do not add work beyond what was required."
    ],
    "postcheck": [
      "The step continues toward completion.",
      "patch_payload differs from local_snippet."
    ]
  },
  "3.2": {
    "category": "3.2",
    "mast_definition": "Weak Verification \u2014 Performs only superficial or incomplete verification.",
    "patch_side_default": "replace_step_content",
    "slot_schema": {
      "MISSED_VERIFICATION": "The verification check that was skipped or performed superficially",
      "THOROUGH_CHECK": "What a thorough verification would include"
    },
    "error_type_spec_text": "error_type: 3.2 Weak Verification\nmast_definition: Performs only superficial or incomplete verification.\npatch_side_default: replace_step_content\nslot_extraction:\n- MISSED_VERIFICATION: identify what verification was skipped or done superficially.\n- THOROUGH_CHECK: describe what a thorough verification would include.\nrepair_instruction:\n- Rewrite the step to include a thorough, substantive verification of the result.\n- The verification must check the actual output against the task requirements.\nforbidden_actions:\n- Do not fabricate verification results.\n- Do not add verification of aspects unrelated to the task requirements.\npostcheck:\n- The step performs thorough verification.\n- patch_payload differs from local_snippet.",
    "repair_instruction": "Rewrite to perform thorough verification against task requirements.",
    "forbidden_actions": [
      "Do not fabricate verification results.",
      "Do not add unrelated verification checks."
    ],
    "postcheck": [
      "Verification is now thorough and substantive.",
      "patch_payload differs from local_snippet."
    ]
  },
  "3.3": {
    "category": "3.3",
    "mast_definition": "No or Incorrect Verification \u2014 Skips verification entirely or verifies against wrong criteria.",
    "patch_side_default": "replace_step_content",
    "slot_schema": {
      "CORRECT_CRITERION": "The correct verification criterion that should have been used",
      "WRONG_OR_MISSING_CHECK": "What verification was missing or wrong"
    },
    "error_type_spec_text": "error_type: 3.3 No or Incorrect Verification\nmast_definition: Skips verification entirely or verifies against wrong criteria.\npatch_side_default: replace_step_content\nslot_extraction:\n- CORRECT_CRITERION: identify the correct criterion the output should have been verified against.\n- WRONG_OR_MISSING_CHECK: identify what verification was missing or incorrectly applied.\nrepair_instruction:\n- Rewrite the step to verify the result against the CORRECT criterion.\n- If no verification was done, add an explicit verification step.\n- If wrong criteria were used, correct them.\nforbidden_actions:\n- Do not fabricate verification outcomes.\n- Do not remove other valid content from the step.\npostcheck:\n- Verification is now present and uses the correct criterion.\n- patch_payload differs from local_snippet.",
    "repair_instruction": "Add or correct verification using the right criterion.",
    "forbidden_actions": [
      "Do not fabricate verification outcomes.",
      "Do not remove other valid content from the step."
    ],
    "postcheck": [
      "Verification is now present and uses the correct criterion.",
      "patch_payload differs from local_snippet."
    ]
  }
}
\end{codebox}

\paragraph{Rule-based postcheck.}
After every patch-LLM call, an independent rule-based postcheck inspects the returned \texttt{patch\_payload} and \texttt{slot\_values}. A patch must pass all applicable checks before its rerun can be scored. Otherwise the generator retries up to \texttt{max\_retries} times before logging the case as a postcheck failure. The universal checks apply to every category. Category-specific checks apply only to entries whose \texttt{patch\_library} category declares them. We list the active checks below. The full specification is reflected in the per-category \texttt{postcheck} arrays inside the JSON listings above.

\begin{itemize}\itemsep0pt
  \item \textbf{Universal.}
    \begin{itemize}\itemsep0pt
      \item \texttt{patch\_payload} is non-empty.
      \item \texttt{patch\_payload} differs from \texttt{local\_snippet} (the patch must change something).
    \end{itemize}
  \item \textbf{Formatting Errors (TRAIL).}
    \begin{itemize}\itemsep0pt
      \item All literal markers listed in \texttt{slot\_values.REQUIRED\_MARKERS} occur exactly in \texttt{patch\_payload}.
      \item No novel \texttt{<...>} tokens are introduced beyond those already present in \texttt{local\_snippet} or in the required-marker set.
    \end{itemize}
  \item \textbf{Tool Selection Errors (TRAIL).}
    \begin{itemize}\itemsep0pt
      \item The string named in \texttt{slot\_values.WRONG\_TOOL}, if any, does not appear in \texttt{patch\_payload}.
      \item The string named in \texttt{slot\_values.CORRECT\_TOOL\_HINT}, if any, appears in \texttt{patch\_payload}.
    \end{itemize}
  \item \textbf{Resource Abuse on remapped \textsc{tool} spans (TRAIL).}
    When the source span was originally annotated as a \textsc{tool} span and was remapped to its parent LLM span for patching (so the local snippet is a \texttt{tool\_calls} JSON object), each tool-call entry in the patched JSON must contain only keys in the OpenAI schema (\texttt{id}, \texttt{type}, \texttt{function}). Novel keys are rejected.
\end{itemize}
The patch generator runs an LLM-side self-check in parallel (the \texttt{postcheck} field in the response schema above), but only the rule-based check gates retries.

\paragraph{Repair verifier (Judge A).}
A separate LLM call decides whether the patch actually removed the source error \(A\). Only patches with \texttt{resolved=True} are forwarded to effect estimation. TRAIL has two user-template variants because its two patch sides probe different invariants (the patched LLM output vs.\ the patched LLM input context). MAST uses a single template over step content.

\begin{promptbox}{Judge A (repair verifier) system prompt (TRAIL and MAST)}
You are verifying whether a source error of type A has been eliminated by a patch.

PRIMARY TASK: Compare ORIGINAL_SPAN with PATCHED_SPAN and determine whether the specific labeled
error A (defined by SOURCE_ERROR_TYPE, ERROR_DESCRIPTION, ERROR_EVIDENCE) is no longer present
in the patched version.

IMPORTANT RULES:
1. Base your verdict on the ORIGINAL_SPAN vs PATCHED_SPAN comparison first.
2. RERUN_SUFFIX is supplementary context only. Do NOT use rerun failures to override a clear
   patch-level fix. If the patched span itself resolves error A, mark resolved=true even if
   the rerun encountered downstream difficulties or repeated errors in OTHER spans.
3. Focus solely on error A. Do not penalize for downstream errors (type B) that the patch
   was not designed to fix.

Return ONLY JSON:
{
  "resolved": true,
  "confidence": 0.0,
  "evidence_excerpt": "string"
}
\end{promptbox}

\begin{promptbox}{Judge A user template --- output-side patch (TRAIL)}
SOURCE_ERROR_TYPE: {A}
ERROR_DESCRIPTION: {ERROR_DESCRIPTION}
ERROR_EVIDENCE: {ERROR_EVIDENCE}

ORIGINAL_SPAN (content that was replaced):
<<<
{ORIGINAL_SNIPPET}
>>>

PATCHED_SPAN (replacement content):
<<<
{PATCH_PAYLOAD}
>>>

{RERUN_CONTEXT_BLOCK}

Task: Has error A been eliminated in the PATCHED_SPAN compared to ORIGINAL_SPAN?
Focus on whether the specific error criterion is met in the patched content itself.
Respond with JSON only: {"resolved": bool, "confidence": float 0-1, "evidence_excerpt": "string"}
\end{promptbox}

\begin{promptbox}{Judge A user template --- input-side patch (TRAIL)}
SOURCE_ERROR_TYPE: {A}
ERROR_DESCRIPTION: {ERROR_DESCRIPTION}
ERROR_EVIDENCE: {ERROR_EVIDENCE}

NOTE: This patch modifies the INPUT (message history context) seen by the LLM, not the LLM's
output. The error A was caused by the context in the original input. The intervention removes
or corrects that error-causing context so the LLM has better information going forward.

ORIGINAL_INPUT (message history before patching):
<<<
{ORIGINAL_SNIPPET}
>>>

PATCHED_INPUT (message history after patching):
<<<
{PATCH_PAYLOAD}
>>>

{RERUN_CONTEXT_BLOCK}

Task: Has the error-causing context for error A been removed or corrected in the PATCHED_INPUT?
Do NOT require the patched input to contain the correct output behavior --- it is an input, not
an output. Check whether the specific pattern described in ERROR_EVIDENCE/ERROR_DESCRIPTION
is absent or corrected in the patched context compared to the original.
Respond with JSON only: {"resolved": bool, "confidence": float 0-1, "evidence_excerpt": "string"}
\end{promptbox}

\begin{promptbox}{Judge A user template --- step content (MAST)}
SOURCE_ERROR_TYPE: {A}
ERROR_DESCRIPTION: {ERROR_DESCRIPTION}
ERROR_EVIDENCE: {ERROR_EVIDENCE}

ORIGINAL_STEP (content that was replaced):
<<<
{ORIGINAL_SNIPPET}
>>>

PATCHED_STEP (replacement content):
<<<
{PATCH_PAYLOAD}
>>>

{RERUN_CONTEXT_BLOCK}

Task: Has error A been eliminated in the PATCHED_STEP compared to ORIGINAL_STEP?
Focus on whether the specific error criterion is met in the patched content itself.
Respond with JSON only: {"resolved": bool, "confidence": float 0-1, "evidence_excerpt": "string"}
\end{promptbox}

\paragraph{Effect evaluator (Judge B).}
For each verified intervention, Judge B inspects the original and counterfactual trace suffixes and assigns an effect label to the downstream category \(B\). The eight effect labels distinguish removal (\texttt{disappeared}), partial reduction (\texttt{delayed}, \texttt{weakened}), no effect (\texttt{unchanged}), negative side effects (\texttt{earlier}, \texttt{strengthened}, \texttt{emerged}), and uninformative cases (\texttt{not\_observable}). Only \texttt{disappeared} contributes a positive count to \(\Delta(A \to B)\). The labels \texttt{emerged} and \texttt{strengthened} contribute negatively, and the remaining labels are conservative neutrals. TRAIL and MAST use the same system prompt and label set, but the surface wording mentions ``spans'' for TRAIL and ``steps in a multi-agent conversation'' for MAST.

\begin{promptbox}{Judge B (effect evaluator) system prompt (TRAIL)}
You are evaluating the downstream effect of a do(A=0) intervention on error type B.

The source error A was locally patched at one labeled span.
The rerun trace suffix shows the counterfactual execution after the intervention.

You must judge ONLY the downstream error type B.

Effect labels:
- disappeared    : B was present in baseline, absent in rerun (intervention removed B)
- delayed        : B was present in baseline, appears later in rerun
- unchanged      : B was present in baseline, appears at similar position in rerun
- earlier        : B was present in baseline, appears earlier in rerun
- weakened       : B was present in baseline, present in rerun but less severe
- strengthened   : B was present in baseline, present in rerun and more severe
- emerged        : B was ABSENT in baseline, but NOW PRESENT in rerun (intervention introduced B)
- not_observable : B was absent in baseline and absent in rerun; effect cannot be assessed

Return ONLY JSON.
\end{promptbox}

\begin{promptbox}{Judge B user template (TRAIL and MAST share this structure)}
SOURCE_ERROR_TYPE: {A}
TARGET_ERROR_TYPE: {B}

TARGET_ERROR_DEFINITION:
{B_TAXONOMY_DEFINITION_OR_INSTANCE_DESCRIPTION}

ORIGINAL_TRACE_SUFFIX:
<<<
{ORIGINAL_SUFFIX}
>>>

ORIGINAL_ONSET_REF:
{ORIGINAL_B_ONSET}

RERUN_TRACE_SUFFIX_AFTER_DO_A_0:
<<<
{RERUN_SUFFIX}
>>>

Task:
Judge how B changed after the do(A=0) intervention.

Required output schema:
{
  "source_error_type": "string",
  "target_error_type": "string",
  "effect_label": "disappeared|delayed|unchanged|earlier|weakened|strengthened|emerged|not_observable",
  "target_present_after": true,
  "original_onset_ref": "string|null",
  "rerun_onset_ref": "string|null",
  "confidence": "high|medium|low",
  "evidence": "string"
}
\end{promptbox}

\paragraph{Edge effect statistic and validation rule.}
For each causal-candidate edge $A\!\to\!B \in \mathcal{R}_C$, Judge~B labels whether the downstream error $B$ is still present in the counterfactual continuation, giving $X_B^{\mathrm{cf}}(T;A)\in\{0,1\}$; the labels \texttt{disappeared}, \texttt{emerged}, \texttt{weakened}, \texttt{strengthened}, and \texttt{not\_observable} are recorded as annotation and reduced to this indicator. The edge-level effect is the Boolean risk-difference of Eq.~\eqref{eq:delta}, averaged over verified interventions
\(\Delta(A \to B)=\tfrac{1}{|\mathcal{T}^{\mathrm{ver}}_{A\prec B}|}\sum_{T\in \mathcal{T}^{\mathrm{ver}}_{A\prec B}} \bigl[X_B(T) - X_B^{\mathrm{cf}}(T;A)\bigr]\),
where \(\mathcal{T}^{\mathrm{ver}}_{A\prec B}\subseteq \mathcal{T}_{A\prec B}\) keeps only traces whose Judge-A verdict marks the source error as resolved. An edge is added to the validated graph $\mathcal{G}_V$ when the number of verified interventions meets a minimum-support requirement and $\Delta(A \to B) > \tau_{\mathrm{val}}$, with $\tau_{\mathrm{val}} = 0.15$.

\paragraph{Intervention pipeline yields.}
Table~\ref{tab:intervention_yields} reports the per-stage success rates of the intervention pipeline for both benchmarks. Rule-based postcheck failures are rare: $\sim$4\% on average, with the worst-case category at 11.1\%. Judge-B effect labels differ qualitatively across benchmarks: TRAIL re-runs the agent live and produces \texttt{disappeared} in 57.5\% of trials, whereas MAST uses LLM-simulated step continuations and is dominated by \texttt{not\_observable} ($\sim$70\%) because the simulator cannot always determine whether the downstream error would have surfaced. After applying the validation rule ($\Delta\!>\!\tau_{\mathrm{val}}$, $n\!\ge\!1$), the candidate-to-validated yield is 12/13 (92.3\%) on TRAIL and 11/23 (47.8\%) on MAST.

\begin{table}[!tbp]
\centering
\footnotesize
\setlength{\tabcolsep}{4pt}
\renewcommand{\arraystretch}{1.05}
\begin{tabular}{l rr}
\toprule
\textbf{Stage / metric} & \textbf{TRAIL} & \textbf{MAST} \\
\midrule
Causal-candidate edges $|\mathcal{R}_C|$                  & 13     & 23     \\
Active $A$-categories with patches                       & 6      & 8      \\
Mean patch-failure rate                                  & 4.1\%  & 4.6\%  \\
\quad worst category                                     & 11.1\% & 11.1\% \\
Total verified interventions                             & 308    & 396    \\
\quad min / median / max per edge                        & 1\,/\,16\,/\,73 & 0\,/\,8\,/\,81 \\
\midrule
Judge-B \texttt{disappeared}                             & 57.5\% & 23\% \\
Judge-B \texttt{not\_observable}                         &  9.9\% & 70\% \\
Judge-B other labels                                     & 32.6\% &  7\% \\
\midrule
Validated edges (after threshold)                        & 12     & 11     \\
Validation rate                                          & 92.3\% & 47.8\% \\
\bottomrule
\end{tabular}
\caption{Per-stage yields of the intervention validation pipeline. Patch-failure rate is averaged over $A$-categories that actually had patches generated. Judge-B percentages on TRAIL are computed over $n{=}294$ effect-label trials across the 13 causal-candidate edges in $\mathcal{R}_C$. MAST percentages are computed over $\sim$340 trials across validated edges and are dominated by \texttt{not\_observable} due to LLM-simulated rerun. Validation rule: $\Delta\!>\!\tau_{\mathrm{val}}$ and $n_{\mathrm{valid}}\!\ge\!1$.}
\label{tab:intervention_yields}
\end{table}

\paragraph{Implementation details.}
Table~\ref{tab:implementation_details} lists the models and thresholds used at each stage of the pipeline. Patch generation, both judges, MAST step-onset annotation, and the MAST simulated rollout use GPT-4o~\cite{openai2024gpt4o}; the TRAIL counterfactual rollout uses o3-mini~\cite{openai2025o3mini}, matching the backbone that generated the original traces. All judge and patch-generation calls use a single model family; the counterfactual rollout differs by benchmark because TRAIL traces can be replayed live whereas MAST traces are recorded multi-agent conversations.

\begin{table}[!tbp]
\centering
\small
\setlength{\tabcolsep}{4pt}
\renewcommand{\arraystretch}{1.1}
\begin{tabular}{lll}
\toprule
\textbf{Component} & \textbf{TRAIL} & \textbf{MAST} \\
\midrule
Patch generation              & GPT-4o & GPT-4o \\
Judge A (repair verification) & GPT-4o & GPT-4o \\
Judge B (effect evaluation)   & GPT-4o & GPT-4o \\
Counterfactual rollout        & o3-mini & GPT-4o \\
\quad rollout mode            & live replay & simulated \\
Step-onset annotation         & gold spans & GPT-4o \\
\midrule
$\tau$ (inference graph)                   & 0.35 & 0.50 \\
$\tau_{\mathrm{val}}$ (edge validation)    & 0.15 & 0.15 \\
$\tau_{\mathrm{GI}}$ (Stage-2 propagation) & 0.10 & 0.10 \\
\bottomrule
\end{tabular}
\caption{Models and thresholds used in the graph-construction and detection pipeline. TRAIL counterfactuals are produced by live agent replay with o3-mini, matching the backbone that generated the original GAIA traces; MAST counterfactuals are LLM-simulated step continuations because the AG2 traces are recorded rather than re-executable.}
\label{tab:implementation_details}
\end{table}

\subsection{Step Annotation Coverage} \label{app:step}

Because MAST provides annotations only at the trace level, we identify the first occurrence of each error category to construct ordered error events. The following table reports the coverage of these step annotations by category and overall.

%

\begin{table}[!tbp]
    \centering
    \small
    \setlength{\tabcolsep}{5pt}
    \renewcommand{\arraystretch}{1.15}
    \begin{tabular}{l l r}
    \toprule
    \textbf{Code} & \textbf{Category} & \textbf{Cov.\,(\%)} \\
    \midrule
    1.1 & Disobey Task Specification        & 76.7 \\
    1.2 & Disobey Role Specification        & 85.7 \\
    1.3 & Step Repetition                   & 95.4 \\
    1.4 & Loss of Conversation History      & 79.5 \\
    1.5 & Unaware of Termination Conditions & 86.1 \\
    \midrule
    2.1 & Conversation Reset                & 75.9 \\
    2.2 & Fail to Ask for Clarification     & 80.7 \\
    2.3 & Task Derailment                   & 78.1 \\
    2.4 & Information Withholding           & 57.1 \\
    2.6 & Action-Reasoning Mismatch         & 70.7 \\
    \midrule
    3.1 & Premature Termination             & 82.8 \\
    3.2 & Weak Verification                 & 96.4 \\
    3.3 & No or Incorrect Verification      & 87.0 \\
    \midrule
    \multicolumn{2}{l}{\textbf{Overall}} & \textbf{82.0} \\
    \bottomrule
    \end{tabular}
    \caption{%
      Step-annotation coverage on the AG2 split ($N{=}393$, GPT-4o annotator).
      \textbf{Cov.} is the fraction of trace-level error instances assigned a step location.
      Overall coverage is 82.0\% (1{,}279\,/\,1{,}560 trace-level instances located).
      False-alarm rate is 0 by pipeline design.%
    }
    \label{tab:step_annotation}
\end{table}

\subsection{Detection Pipeline and +CG Ablation} \label{app:detection_pipeline}

This appendix collects the formal Stage-2 prediction recipe used by the graph-guided detector of Section~\ref{graph_guided_detection}, the definition of the naive single-stage Static Graph Guidance (+CG) baseline, and the analysis of why +CG underperforms. The empirical comparison itself is reported in Table~\ref{tab:cg_main_results}.

\paragraph{Stage-2 prediction.}
Given the Stage-1 prediction $\hat{E}^{(1)}(T)$ and the trace-specific edge subset $\mathcal{R}_\tau^{(T)}$, the Stage-2 prediction is
\[
\hat{E}^{(2)}(T) = \mathcal{J}^{(2)}(T,\mathcal{C},\hat{E}^{(1)}(T),\mathcal{R}_\tau^{(T)}),
\]
where $\mathcal{J}^{(2)}$ is the second-stage judge call, instantiated with a targeted prompt that conditions on $\hat{E}^{(1)}(T)$ as already-detected errors and on $\mathcal{R}_\tau^{(T)}$ as trace-specific guidance.

\paragraph{Final \our prediction.}
The two stages are merged with category-level deduplication against Stage~1:
\[
\begin{aligned}
\hat{E}^{\our}(T) ={} & \hat{E}^{(1)}(T) \\
 & {}\cup \{(c,\ell)\in\hat{E}^{(2)}(T): c\notin D(T)\}.
\end{aligned}
\]

\paragraph{Static Graph Guidance (+CG): method.}
The simplest way to inject \(\mathcal{G}_\tau\) is to format its full edge set $\mathcal{R}_\tau$ as a guidance block and prepend it to the baseline prompt. For the validated-only variant ($\mathcal{G}_V$) the block describes intervention-validated causal effects. For the thresholded union variant ($\mathcal{G}_\tau$) it additionally describes correlation patterns from $\mathcal{G}_S$ anchored by validated causal edges. The single-stage prediction uses the same judge $\mathcal{J}$ and edge set $\mathcal{R}_\tau$ as Stage-1 of our method (\S\ref{graph_guided_detection}):
\[
\hat{E}^{+\mathrm{CG}}(T) = \mathcal{J}(T,\mathcal{C},\mathcal{R}_\tau).
\]
The difference from \our is operational rather than notational. The same edge set is supplied as a static, trace-agnostic context block in a single LLM call, with no trace-specific filtering and no Stage-2 verification.

\paragraph{Why +CG underperforms \our.}
+CG demonstrates that \emph{graph access alone is not the source of \our's gain}. The full inference graph is identical in both methods. The difference is how it is consumed. +CG presents every edge in $\mathcal{G}_\tau$ as static context, so edges with sources absent from the trace act as distractors during Stage-1 reasoning, and there is no Stage-2 verification to discharge or confirm propagated hypotheses. On long-context inputs (TRAIL-SWE-Bench, with median trace $\approx$ 213\,K characters and tail traces exceeding 1\,M), the static prepend additionally consumes context budget that the trace needs, and a non-trivial fraction of samples exceed the model's max input length once the guidance block is added---degrading even raw Stage-1 coverage. \our instead restricts Stage-2 to the trace-specific edge subset $\mathcal{R}_\tau^{(T)}$ derived from the Stage-1 detection profile (Section~\ref{graph_guided_detection}), so only edges with a detected upstream source are surfaced to the judge, each propagated downstream hypothesis is verified against the trace before being merged, and the per-call payload stays bounded. The two-stage design is therefore the operational lever. The inference graph supplies the search space, but conditioning that space on the Stage-1 profile is what turns it into a useful prior. Per-backbone Stage-2 trigger rates are reported in Appendix~\ref{app:results_details}.

\subsection{Threshold and Stage-2 Diagnostics} \label{app:results_details}

This appendix consolidates two diagnostics that characterize the inference graph $\mathcal{G}_\tau$ used in the main results: a graph-richness sensitivity sweep over the threshold $\tau$, and per-(benchmark, backbone) Stage-2 trigger rates.

\paragraph{Threshold and graph-richness sensitivity.}
We analyze the inference-graph threshold \(\tau\) as a graph-richness parameter. For each candidate threshold, we construct the corresponding thresholded union graph \(\mathcal{G}_\tau\), run the graph-guided detector across evaluated backbones, and compute model-averaged weighted-F1. We report one benchmark-level setting in the main table, \(\tau=0.35\) for TRAIL and \(\tau=0.50\) for MAST. This analysis is conducted on the benchmark corpus and is intended as sensitivity analysis rather than held-out hyperparameter tuning. To avoid model-specific overfitting, we use a single \(\tau\) per benchmark rather than tuning \(\tau\) separately for each backbone.

\paragraph{Score and variants.}
The corr-screening score is the Suppes geomean
$w_S(A,B)=\sqrt{P(A\!\prec\!B\mid \text{both})\cdot\Delta_{\mathrm{PR}}(A\!\to\!B)}$,
evaluated on Suppes-screened pairs. For each candidate $\tau$, $\mathcal{G}_\tau$ is the union of (i) all observational edges with $w_S(A,B)\geq\tau$ and (ii) the intervention-validated edges $\mathcal{R}_V$ (kept regardless of $w_S$). The sweep in Tables~\ref{tab:threshold_sweep_ablation} and~\ref{tab:mast_ablation_threshold_sweep} compares this to two controls:
\textbf{causal-only} keeps only $\mathcal{R}_V$ (no observational extension);
\textbf{random-$K$} replaces $\mathcal{R}_V$ with $K$ directed edges sampled uniformly (seed $42$) from category pairs not in the Suppes screen, matched in edge count to causal-only.

%

\begin{table*}[!tbp]
\centering
\small
\setlength{\tabcolsep}{5pt}
\renewcommand{\arraystretch}{1.2}
\begin{tabular}{l l rrr @{\hspace{10pt}} rrr}
\toprule
\multirow{2}{*}{\textbf{Model}} &
\multirow{2}{*}{\textbf{Graph Variant (\#edges)}} &
\multicolumn{3}{c}{\textbf{TRAIL-GAIA}} &
\multicolumn{3}{c}{\textbf{TRAIL-SWE-Bench}} \\
\cmidrule(lr){3-5}\cmidrule(lr){6-8}
& & F1 & Loc & Joint & F1 & Loc & Joint \\
\midrule
\multirow{5}{*}{Mistral-Small-3.1-24B}
  & random-12        (12) & 25.57 & 22.61 &  9.79 &  9.00 &  3.56 &  0.00 \\
  & causal-only      (12) & 30.76 & 27.79 & 11.51 & 12.45 &  4.72 &  0.00 \\
  & corr~$\geq$~0.35 (19) & 34.15 & 25.71 & \textbf{12.27} & \textbf{14.40} & \textbf{7.59} & \textbf{0.87} \\
  & corr~$\geq$~0.25 (21) & \textbf{34.76} & 28.13 & 10.68 &  9.27 &  3.92 &  0.00 \\
  & corr~$\geq$~0.20 (25) & 33.99 & \textbf{28.15} & 11.05 &  7.86 &  5.42 &  0.25 \\[3pt]
\multirow{5}{*}{GPT-oss-120B}
  & random-12        (12) & 30.57 & 21.72 &  6.93 & \textbf{35.30} & \textbf{3.91} & \textbf{0.31} \\
  & causal-only      (12) & 30.14 & 24.47 &  6.34 & 27.13 &  2.92 &  0.00 \\
  & corr~$\geq$~0.35 (19) & 37.20 & \textbf{28.00} &  9.48 & 30.07 &  1.60 &  0.00 \\
  & corr~$\geq$~0.25 (21) & 38.69 & 21.96 &  7.83 & 35.02 &  1.02 &  0.25 \\
  & corr~$\geq$~0.20 (25) & \textbf{40.62} & 25.99 & \textbf{12.81} & 27.72 &  1.25 &  0.00 \\[3pt]
\multirow{5}{*}{GPT-oss-20B}
  & random-12        (12) & 25.17 & 12.12 &  4.26 & 23.50 &  1.11 &  0.00 \\
  & causal-only      (12) & 22.29 & \textbf{18.29} &  4.40 & 13.03 &  1.33 &  0.42 \\
  & corr~$\geq$~0.35 (19) & \textbf{33.28} & 12.28 &  4.09 & \textbf{29.89} &  1.23 & \textbf{0.43} \\
  & corr~$\geq$~0.25 (21) & 33.11 & 12.59 &  5.02 & 27.71 & \textbf{1.78} &  0.28 \\
  & corr~$\geq$~0.20 (25) & 30.30 & 13.27 & \textbf{5.54} & 21.13 &  0.42 &  0.00 \\[3pt]
\multirow{5}{*}{Gemma-3-27B-IT}
  & random-12        (12) & 21.46 &  7.27 &  0.20 & 15.09 &  1.47 &  0.00 \\
  & causal-only      (12) & \textbf{25.30} & 11.00 & \textbf{1.33} & 15.16 & \textbf{2.36} &  0.00 \\
  & corr~$\geq$~0.35 (19) & 21.26 & \textbf{11.20} &  0.73 & 15.41 &  1.47 &  0.00 \\
  & corr~$\geq$~0.25 (21) & 18.96 &  6.87 &  0.27 & 16.79 &  1.47 &  0.00 \\
  & corr~$\geq$~0.20 (25) & 21.12 & 11.00 &  1.13 & \textbf{18.52} &  1.88 &  0.00 \\[3pt]
\multirow{5}{*}{QwenLong-L1-32B}
  & random-12        (12) & 19.57 & 12.20 &  4.08 & 10.10 &  0.31 &  0.00 \\
  & causal-only      (12) & 16.83 & \textbf{19.23} &  3.63 & 11.09 &  0.42 &  0.00 \\
  & corr~$\geq$~0.35 (19) & \textbf{26.57} & 17.32 &  4.13 & 10.33 &  0.42 &  0.00 \\
  & corr~$\geq$~0.25 (21) & 20.75 & 12.43 &  3.05 & \textbf{16.53} & \textbf{0.83} &  0.00 \\
  & corr~$\geq$~0.20 (25) & 25.46 & 17.16 & \textbf{5.97} & 13.20 &  0.42 &  0.00 \\
\bottomrule
\end{tabular}
\caption{%
  Graph-richness ablation on TRAIL under the two-pass dynamic injector (+GI). Variants are defined in \S\ref{app:results_details} (\textbf{random-12}, \textbf{causal-only}, \textbf{corr~${\geq}\tau$} for $\tau\in\{0.35,0.25,0.20\}$ giving sizes $\{19,21,25\}$). Open-source panel only; closed-source Gemini models excluded. \textbf{Bold}: best per model and metric column. All metrics in \%.%
}
\label{tab:threshold_sweep_ablation}
\end{table*}

\begin{table*}[!tbp]
  \centering
  \small
  \setlength{\tabcolsep}{6pt}
  \renewcommand{\arraystretch}{1.2}
  \begin{tabular}{l l rr}
  \toprule
  \textbf{Model} & \textbf{Graph Variant (\#edges)} & \textbf{F1} & \textbf{Macro-F1} \\
  \midrule
  \multirow{5}{*}{Mistral-Small-3.1-24B}
    & random-11        (11) & 36.88              & 25.92              \\
    & causal-only      (11) & 37.01              & 25.86              \\
    & corr~$\geq$~0.60 (18) & 36.95              & 25.97              \\
    & corr~$\geq$~0.50 (25) & \textbf{38.29}     & \textbf{26.97}     \\
    & corr~$\geq$~0.40 (29) & 37.40              & 26.30              \\[3pt]
  \multirow{5}{*}{GPT-oss-120B}
    & random-11        (11) & 25.15              & 18.72              \\
    & causal-only      (11) & 25.26              & 18.50              \\
    & corr~$\geq$~0.60 (18) & 24.72              & 18.18              \\
    & corr~$\geq$~0.50 (25) & 27.21              & 19.63              \\
    & corr~$\geq$~0.40 (29) & \textbf{27.95}     & \textbf{20.55}     \\[3pt]
  \multirow{5}{*}{GPT-oss-20B}
    & random-11        (11) & 21.05              & \textbf{16.33}     \\
    & causal-only      (11) & 21.68              & 16.04              \\
    & corr~$\geq$~0.60 (18) & 20.35              & 15.29              \\
  & corr~$\geq$~0.50 (25) & \textbf{22.34}     & 15.94              \\
    & corr~$\geq$~0.40 (29) & 20.76              & 14.66              \\[3pt]
  \multirow{5}{*}{Gemma-3-27B-IT}
    & random-11        (11) & 16.95              & 13.11              \\
    & causal-only      (11) & 20.25              & 14.02              \\
    & corr~$\geq$~0.60 (18) & 19.39              & 13.54              \\
  & corr~$\geq$~0.50 (25) & \textbf{21.17}     & \textbf{14.56}     \\
    & corr~$\geq$~0.40 (29) & 20.52              & 13.70              \\[3pt]
  \multirow{5}{*}{QwQ-32B$^{*}$}
  & random-11        (11) & \textbf{16.08}     & \textbf{12.69}     \\
    & causal-only      (11) & 13.69              &  9.80              \\
    & corr~$\geq$~0.60 (18) & 15.57              & 11.43              \\
    & corr~$\geq$~0.50 (25) & 15.61              & 11.47              \\
    & corr~$\geq$~0.40 (29) & 16.01              & 11.75              \\
  \bottomrule
  \end{tabular}
  \caption{%
    Graph-richness ablation on MAST ($N{=}393$ AG2 traces) under the two-pass dynamic injector (+GI). Variants are defined in \S\ref{app:results_details} (\textbf{random-11}, \textbf{causal-only}, \textbf{corr~${\geq}\tau$} for $\tau\in\{0.60,0.50,0.40\}$ giving union sizes $\{18,25,29\}$). Open-source panel only; closed-source GPT-4o excluded. \textbf{Bold}: best per model and metric column. All metrics in \%. $^{*}$QwQ-32B uses \texttt{enable\_thinking}.%
  }
  \label{tab:mast_ablation_threshold_sweep}
  \end{table*}

\paragraph{Stage-2 trigger rates.}
We report, for every (benchmark, backbone) cell in the main results table, the fraction of evaluation traces on which \our actually issued a Stage-2 call (Table~\ref{tab:pass2_trigger_rate}). Stage~2 fires only when the trace-specific edge subset $\mathcal{R}_\tau^{(T)}$ is non-empty after the Stage-1 detector profile is propagated through the inference graph $\mathcal{G}_\tau$ (Section~\ref{graph_guided_detection}). A low trigger rate at fixed $\tau$ means Stage~1 already detected enough source categories that no further downstream hypotheses cleared the propagation threshold. A high rate means Stage~1 left targets the graph could still suggest. The trigger rates therefore complement F1 by characterizing how often the second stage is being exercised. Rates are read from the per-trace meta logs written by the eval pipeline (TRAIL: \texttt{\_meta\_<trace\_id>.json}, MAST: per-trace JSON with field \texttt{pass2\_triggered}).

\begin{table}[!tbp]
\centering
\footnotesize
\setlength{\tabcolsep}{4pt}
\renewcommand{\arraystretch}{1.05}
\begin{tabular}{l rr @{\hspace{6pt}} r}
\toprule
\multirow{2}{*}{\textbf{Backbone}} &
\multicolumn{2}{c}{\textbf{TRAIL}} &
\multicolumn{1}{c}{\textbf{MAST}} \\
\cmidrule(lr){2-3} \cmidrule(lr){4-4}
& GAIA & SWE-Bench & all traces \\
\midrule
Mistral-Small-3.1-24B & 54.7\% & 38.7\% & 94.4\% \\
GPT-oss-120B          & 74.4\% & 58.1\% & 51.9\% \\
GPT-oss-20B           & 55.6\% & 58.1\% & 38.4\% \\
Gemma-3-27B-IT        & 70.9\% & 51.6\% & 56.5\% \\
Qwen Family$^{*}$         & 51.3\% & 45.2\% & 21.6\% \\
\midrule
Closed-source$^{*}$               & 59.8\% & 83.9\% & 70.7\% \\
\midrule
Traces evaluated ($n$)            & 117    & 31     & 393    \\
\bottomrule
\end{tabular}
\caption{Stage-2 trigger rate per (benchmark, backbone). A trace contributes one count to the numerator when its Stage-1 prediction profile, propagated through $\mathcal{G}_\tau$, exposes at least one downstream hypothesis above the graph-injection threshold $\tau_{\mathrm{GI}}$. Otherwise Stage~2 is skipped and only Stage-1 predictions are kept. Sample sizes match the main results table. \textbf{\our}: thresholded inference graph (TRAIL $\tau{=}0.35$ with 19 edges, MAST $\tau{=}0.50$ with 25 edges). $^{*}$\textbf{Qwen Family} = QwenLong-L1-32B on TRAIL + QwQ-32B on MAST. \textbf{Closed-source} = Gemini-2.5-Pro on TRAIL + GPT-4o on MAST (same row pairing as Table~\ref{tab:main_results}).}
\label{tab:pass2_trigger_rate}
\end{table}

\subsection{Additional Backbones} \label{app:additional_results}

To test whether graph guidance depends on the generation of the underlying model, we additionally evaluate GPT-5 as a proprietary frontier model and Qwen3.6-35B-A3B as an open-weight model. Both use the same graphs, thresholds, prompts, and scorers as Table~\ref{tab:main_results}, with no re-tuning. Table~\ref{tab:new_models} reports the results. \our improves weighted-F1 in all six model-benchmark cells, including on the strongest proprietary backbone, and the small MAST accuracy decrease on both models follows the accuracy--recall trade-off discussed in Section~\ref{sec:main_results}.


\begin{table*}[!t]
\centering \footnotesize
\setlength{\tabcolsep}{3.5pt}
\renewcommand{\arraystretch}{1.0}
\begin{tabular}{l l rrr @{\hspace{6pt}} rrr @{\hspace{6pt}} rrrr}
\toprule
\multirow{2}{*}{\textbf{Model}} & \multirow{2}{*}{\textbf{Method}} & \multicolumn{3}{c}{\textbf{TRAIL-GAIA}} & \multicolumn{3}{c}{\textbf{TRAIL-SWE-Bench}} & \multicolumn{4}{c}{\textbf{MAST}} \\
\cmidrule(lr){3-5} \cmidrule(lr){6-8} \cmidrule(lr){9-12}
& & F1 & Loc & Joint & F1 & Loc & Joint & F1 & Precision & Recall & Acc \\
\midrule
\multirow{3}{*}{GPT-5}
  & Baseline    & 50.82 & 58.34 & 27.81 & 43.11 & 5.25 & 1.41 & 31.49 & 27.63 & 19.58 & 66.18 \\
  & \our        & 53.00 & 58.92 & 36.10 & 63.29 & 10.81 & 1.56 & 35.92 & 29.17 & 26.19 & 64.22 \\
  & $\Delta$     & \dup{+2.17} & \dup{+0.58} & \dup{+8.29} & \dup{+20.19} & \dup{+5.55} & \dup{+0.15} & \dup{+4.43} & \dup{+1.54} & \dup{+6.61} & \ddn{-1.96} \\
\multirow{3}{*}{Qwen3.6-35B-A3B}
  & Baseline    & 21.50 & 16.68 & 4.78 & 12.33 & 0.00 & 0.00 & 24.11 & 29.65 & 16.39 & 66.51 \\
  & \our        & 24.88 & 14.63 & 5.42 & 20.86 & 0.63 & 0.00 & 29.76 & 29.68 & 21.01 & 64.77 \\
  & $\Delta$     & \dup{+3.38} & \ddn{-2.05} & \dup{+0.64} & \dup{+8.53} & \dup{+0.63} & 0.00 & \dup{+5.65} & \dup{+0.03} & \dup{+4.62} & \ddn{-1.74} \\
\bottomrule
\end{tabular}
\caption{Additional backbones. Same columns and metrics as Table~\ref{tab:main_results}: weighted-F1 for TRAIL and MAST, Loc/Joint for TRAIL, macro Precision/Recall/Acc for MAST. \textbf{\our}: thresholded inference graph (TRAIL $\tau{=}0.35$; MAST $\tau{=}0.50$) with two-stage dynamic injection, identical to the main table. All metrics in \%.}
\label{tab:new_models}
\end{table*}

\subsection{Who\&When-Style Prompting: Adaptation and Full Results}
\label{app:ww_adaptation}

Who\&When formulates failure attribution as identifying the agent and step responsible for a failed multi-agent execution. Its original prompting strategies are designed for a single decisive error, whereas our benchmarks require multi-error prediction. We therefore adapt the prompting format while preserving the intended inference style of each strategy. This appendix collects both the adaptation details and the per-cell tables that back Figure~\ref{fig:ww_bars} in \S\ref{sec:ww}.

\paragraph{All-at-once.}
The all-at-once variant receives the full trace and the benchmark taxonomy in a single call. Instead of asking for one responsible error, we ask the judge to output all supported error categories. For TRAIL, the output contains category and span identifier pairs. For MAST, the output is a trace-level yes/no vector over the taxonomy. The graph-guided version prepends the selected EDGE inference graph to the same prompt.

\paragraph{Step-by-step.}
The step-by-step variant evaluates the trace through an exhaustive stepwise scan. At each step, the judge receives the local step context, the taxonomy, and the accumulated prediction state. Predictions are aggregated across steps to form the final trace-level output. For TRAIL, each predicted category is paired with the corresponding span or step identifier. For MAST, step-level predictions are OR-aggregated into a trace-level multi-label vector. The graph-guided version uses the same stepwise protocol, with the selected EDGE inference graph included as additional guidance.

\paragraph{Binary search.}
We do not include the binary-search variant in the main comparison. Binary search is useful when attribution seeks a single decisive point because recursive interval narrowing reduces the number of inspected steps. In the multi-error setting, however, multiple categories may emerge at different points in the same trace. Attributing all errors would require maintaining and recursively expanding several candidate intervals, which removes the efficiency advantage and makes the adaptation less faithful to the original strategy. We therefore report all-at-once and step-by-step as the two representative prompting regimes: one holistic and one exhaustive.

\paragraph{Graph-guided variants.}
For both adapted strategies, the +Graph condition uses the same selected EDGE inference graph as the main experiments. TRAIL uses \(\tau=0.35\), and MAST uses \(\tau=0.50\). The graph is fixed after development. It is not updated during evaluation.

\paragraph{Full per-cell tables.}
Table~\ref{tab:who_and_when_results} reports the per-cell TRAIL numbers (the data visualized in Figure~\ref{fig:ww_bars}) including the Joint column that is not shown in the figure. Table~\ref{tab:mast_who_and_when} reports MAST, which is tabular-only because the main paper only visualizes the TRAIL splits.

%

\begin{table*}[!tbp]
\centering
\small
\setlength{\tabcolsep}{5pt}
\renewcommand{\arraystretch}{1.2}
\begin{tabular}{l l rrr @{\hspace{10pt}} rrr}
\toprule
\multirow{2}{*}{\textbf{Model}} &
\multirow{2}{*}{\textbf{Method}} &
\multicolumn{3}{c}{\textbf{TRAIL-GAIA}} &
\multicolumn{3}{c}{\textbf{TRAIL-SWE-Bench}} \\
\cmidrule(lr){3-5}\cmidrule(lr){6-8}
& & F1 & Loc & Joint & F1 & Loc & Joint \\
\midrule
\multirow{4}{*}{Mistral-Small-3.1-24B}
  & All-at-once            & 22.41 & 27.59 &  4.20 &
                   24.81 &  6.78 &  0.58 \\
  & Step-by-step            & 14.74 & 25.06 &  2.99 &
                    3.93 &  1.89 &  0.28 \\
  & All-at-once + Graph      & \textbf{29.35} & \textbf{30.86} & \textbf{ 5.93} &
                   \textbf{25.04} & 16.91 &  3.92 \\
  & Step-by-step + Graph      & 15.61 & 20.93 &  4.21 &
                   19.41 & \textbf{26.64} & \textbf{5.77} \\[3pt]
\multirow{4}{*}{GPT-oss-120B}
  & All-at-once            & 15.87 & 27.19 &  3.81 &
                   19.10 &  5.35 &  0.56 \\
  & Step-by-step            & 22.43 & 45.56 &  1.94 &
                   13.29 & 26.48 &  0.88 \\
  & All-at-once + Graph      & 33.11 & 32.63 & \textbf{12.54} &
                   \textbf{41.07} &  6.27 &  1.35 \\
  & Step-by-step + Graph      & \textbf{35.02} & \textbf{59.28} & 10.62 &
                   35.81 & \textbf{52.35} & \textbf{9.24} \\[3pt]
\multirow{4}{*}{GPT-oss-20B}
  & All-at-once            & 13.88 & 23.72 &  3.59 &
                    8.32 &  1.61 &  0.00 \\
  & Step-by-step            & 18.73 & \textbf{28.59} &  3.15 &
                   18.19 & 18.22 &  0.25 \\
  & All-at-once + Graph      & \textbf{29.34} & 23.17 & \textbf{ 6.25} &
                   11.76 &  2.50 &  0.00 \\
  & Step-by-step + Graph      & 13.71 & 16.83 &  2.07 &
                   \textbf{22.91} & \textbf{21.63} & \textbf{4.43} \\[3pt]
\multirow{4}{*}{Gemma-3-27B-IT}
  & All-at-once            & 14.75 & 12.61 &  2.69 &
                   12.90 &  6.41 &  0.25 \\
  & Step-by-step            & 20.18 & 22.26 &  0.97 &
                   10.49 & 11.72 &  0.31 \\
  & All-at-once + Graph      & \textbf{31.56} & 27.50 &  5.75 &
                   \textbf{25.46} &  8.09 &  0.13 \\
  & Step-by-step + Graph      & 30.08 & \textbf{32.50} & \textbf{ 6.71} &
                   23.31 & \textbf{18.68} & \textbf{2.63} \\[3pt]
\multirow{4}{*}{QwenLong-L1-32B}
  & All-at-once            & 11.07 & 16.10 &  1.03 &
                    5.75 &  3.80 &  0.28 \\
  & Step-by-step            & 17.10 & 18.42 &  2.98 &
                    0.00 & 31.63 &  0.00 \\
  & All-at-once + Graph      & \textbf{23.83} & 18.34 & \textbf{ 3.47} &
                   \textbf{17.44} &  0.92 &  0.50 \\
  & Step-by-step + Graph      & 17.69 & \textbf{59.61} &  2.33 &
                   15.00 & \textbf{37.64} & \textbf{ 1.29} \\
\bottomrule
\end{tabular}
\caption{%
  Who\&When prompts applied to TRAIL with and without the
  causal-graph context, on open-source models only.
  \textbf{All-at-once}: the original Who\&When all-at-once strategy adapted
  to multi-error TRAIL attribution.
  \textbf{Step-by-step}: the original Who\&When step-by-step strategy adapted
  with cumulative prefixes.
  \textbf{+ Graph}: the selected \our inference graph is added to the
  corresponding prompting strategy.
  All metrics in \%.
  \textbf{Bold}: best result in each model block and metric column.%
}
\label{tab:who_and_when_results}
\end{table*}

%
%
%

\begin{table*}[!tbp]
\centering
\small
\setlength{\tabcolsep}{8pt}
\renewcommand{\arraystretch}{1.2}
\begin{tabular}{l l rr}
\toprule
\textbf{Model} & \textbf{Method} & \textbf{F1} & \textbf{Macro-F1} \\
\midrule
\multirow{4}{*}{Mistral-Small-3.1-24B}
  & All-at-once     & 23.08              & 15.86              \\
  & Step-by-step     &  5.89              &  3.82              \\
  & All-at-once + Graph  & \textbf{27.13}     & \textbf{21.54}     \\
  & Step-by-step + Graph  &  6.90              &  4.71              \\[3pt]
\multirow{4}{*}{GPT-oss-120B}
  & All-at-once     & 22.85              & 16.14              \\
  & Step-by-step     & 29.44              & 21.29              \\
  & All-at-once + Graph  & 28.00              & 20.44              \\
  & Step-by-step + Graph  & \textbf{31.78}     & \textbf{22.85}     \\[3pt]
\multirow{4}{*}{GPT-oss-20B}
  & All-at-once     & 16.36              & 11.31              \\
  & Step-by-step     & 26.55              & 19.15              \\
  & All-at-once + Graph  & 21.65              & 15.70              \\
  & Step-by-step + Graph  & \textbf{27.27}     & \textbf{19.87}     \\[3pt]
\multirow{4}{*}{Gemma-3-27B-IT}
  & All-at-once     & 18.13              & 11.62              \\
  & Step-by-step     & 11.84              &  8.65              \\
  & All-at-once + Graph  & \textbf{28.13}     & \textbf{20.61}     \\
  & Step-by-step + Graph  & 13.47              &  9.69              \\[3pt]
\multirow{4}{*}{QwQ-32B$^{*}$}
  & All-at-once     & 16.96              & 12.72              \\
  & Step-by-step     & 18.28              & 12.84              \\
  & All-at-once + Graph  & \textbf{24.55}     & \textbf{17.42}     \\
  & Step-by-step + Graph  & 17.65              & 12.02              \\
\bottomrule
\end{tabular}
\caption{%
  Who\&When adaptation results on MAST (393 AG2 traces, all metrics in \%).
  \textbf{All-at-once}: Yin et al.'s all-at-once attribution adapted to a
  trace-level 13-bit yes/no vector (one call per trace).
  \textbf{Step-by-step}: step-by-step attribution OR-aggregated over steps
  ($N$ calls per trace).
  \textbf{+ Graph}: the selected \our inference graph
  (corr~$\geq$~0.50, 25 edges) is added to the corresponding prompting strategy.
  \textbf{Bold}: best result in each model block (for each metric column).%
}
\label{tab:mast_who_and_when}
\end{table*}

\subsection{Error Attribution Analysis Details} \label{app:case_study}

This appendix supplements the error-attribution analysis of Section~\ref{sec:case_study} with per-model per-parent recall numbers (Table~\ref{tab:per_parent_recall}) and a per-trace case-study breakdown (Table~\ref{tab:trail_case_study_gpt120b}).

\paragraph{Per-parent recall numbers.}
Table~\ref{tab:per_parent_recall} reports baseline and \our\ recall for each TRAIL parent category on TRAIL-GAIA, for the three backbones in Figure~\ref{fig:error_attribution_4bucket}. Recall is the fraction of gold errors in that parent that the method correctly attributed.

\begin{table*}[!tbp]
\centering
\footnotesize
\setlength{\tabcolsep}{4pt}
\renewcommand{\arraystretch}{1.05}
\begin{tabular}{l l rrr}
\toprule
\textbf{Model} & \textbf{Method} & \textbf{Reasoning} & \textbf{Sys.\ Exec.} & \textbf{Planning} \\
\midrule
\multirow{3}{*}{GPT-oss-120B}
  & Baseline    & 20.1 & 30.7 &  9.0 \\
  & \our        & 28.4 & 32.5 & 46.6 \\
  & $\Delta$    & \dup{+8.3}  & \dup{+1.8}  & \dup{+37.6} \\[2pt]
\multirow{3}{*}{Mistral-Small-3.1-24B}
  & Baseline    & 15.4 & 28.9 &  7.4 \\
  & \our        & 33.7 & 23.2 & 31.2 \\
  & $\Delta$    & \dup{+18.3} & \ddn{-5.7}  & \dup{+23.8} \\[2pt]
\multirow{3}{*}{Gemini-2.5-Pro}
  & Baseline    & 34.3 & 46.1 & 26.5 \\
  & \our        & 39.1 & 42.1 & 50.8 \\
  & $\Delta$    & \dup{+4.7}  & \ddn{-3.9}  & \dup{+24.3} \\
\bottomrule
\end{tabular}
\caption{Per-parent recall on TRAIL-GAIA ($n{=}638$ gold-error instances; Reasoning $n{=}169$, System Execution $n{=}280$, Planning \& Coord.\ $n{=}189$). Recall is the percentage of gold errors in each parent category that the method correctly attributed. $\Delta = \our - \text{Baseline}$. All metrics in \%.}
\label{tab:per_parent_recall}
\end{table*}

\paragraph{Per-trace case-study breakdown.}
Table~\ref{tab:trail_case_study_gpt120b} reports a four-case GPT-oss-120B breakdown covering both TRAIL splits: two SWE traces (\texttt{c104d0\ldots} working and \texttt{72822d\ldots} failure) and two GAIA traces (Working-1 \texttt{dbc070\ldots} and Not-working-1 \texttt{ea313e\ldots}), so successful and failure regimes are represented on both splits.

\begin{table*}[t]
\centering
\small
\setlength{\tabcolsep}{3pt}
\renewcommand{\arraystretch}{1.12}
\begin{tabular}{p{0.09\linewidth} p{0.06\linewidth} p{0.18\linewidth} p{0.17\linewidth} p{0.25\linewidth} p{0.17\linewidth}}
\toprule
\textbf{Case} &
\textbf{Split} &
\textbf{Trace} &
\textbf{$\Delta$ metrics} &
\textbf{Propagation evidence} &
\textbf{Takeaway} \\
\midrule
Working-1 &
GAIA &
\texttt{dbc070...}, Phys.org, target ``Bravo'' &
F1 $-0.257$; Loc $+1.000$; Joint $+0.667$ &
Pass-1 detects upstream errors; Pass-2 uses 3 filtered edges and adds 2 downstream errors. &
Improves localization and joint alignment but lowers category F1. \\
\midrule
Working-2 &
SWE &
\texttt{c104d0...}, SQLFluff issue fix &
F1 $+0.071$; Loc $+0.333$; Joint $0.000$ &
Pass-1 detects instruction/resource errors; Pass-2 uses 3 edges and adds 2 errors. &
Modest category gain with clearer location gain. \\
\midrule
Not-working-1 &
GAIA &
\texttt{ea313e...}, Doctor Who location task &
F1 $+0.400$; Loc $0.000$; Joint $0.000$ &
Pass-1/2 are active, but augmentation shifts location/category alignment. &
Category gain does not improve span-level attribution. \\
\midrule
Not-working-2 &
SWE &
\texttt{72822d...}, UnicodeEncodeError/file wipe &
F1 $0.000$; Loc $0.000$; Joint $0.000$ &
Pass-1 detects retrieval/output/identification issues; Pass-2 uses 3 edges and adds 2 errors. &
Active propagation gives no measurable gain. \\
\bottomrule
\end{tabular}
\caption{%
  Case study on \textbf{GPT-oss-120B} showing successful and failure regimes of
  \textbf{causal error propagation with corr-thresholded union augmentation}.
  Deltas are \textbf{\our minus baseline}.
  ``Propagation evidence'' reports whether Pass-2 was triggered and how many filtered edges/new errors were introduced.
  The causal-only graph variant is discussed as diagnostic context in the text.%
}
\label{tab:trail_case_study_gpt120b}
\end{table*}

\subsection{Dataset, Compute, and Software} \label{app:setup_resources}

This appendix collects the dataset documentation, inference compute budget, and software-package details that support the experimental setup of Section~\ref{sec:main_results}, in line with the EMNLP Responsible NLP Checklist (B5, C1, C4).

\paragraph{Source benchmarks, language, and license.}
\textbf{TRAIL}~\cite{deshpande2025trail} provides span-level multi-error annotations over 148 agent traces drawn from two upstream task corpora: GAIA~\cite{mialon2024gaia} (general-purpose agentic tasks) and SWE-Bench~\cite{jimenez2024swe} (real GitHub-issue code repair). \textbf{MAST}~\cite{cemri2026multi} provides trace-level multi-label annotations over 393 AutoGen (AG2) multi-agent conversation traces, released as the MAD dataset on Hugging Face (\texttt{mcemri/MAD}). Both corpora are English-only. All trace content and annotations are produced from publicly available agent executions and contain no personally identifiable human data or human-demographic attributes. The TRAIL repository is released under an MIT-style license, and MAD is distributed under the Hugging Face dataset license documented on the dataset card.

\paragraph{Trace counts and taxonomy size.}
Table~\ref{tab:dataset_overview} summarizes the two benchmarks. TRAIL covers 19 leaf categories spanning three top-level groups (Reasoning, Execution, Planning and Coordination). The per-split category histograms are listed in the original TRAIL paper. MAST covers 13 leaf categories spanning three top-level groups (Specification, Inter-Agent Misalignment, Verification), with the per-category step-annotation coverage reported in Table~\ref{tab:step_annotation}. TRAIL contains 841 annotated error instances (585 on GAIA, 256 on SWE-Bench), and MAST contains 1{,}560 positive category labels. The raw traces total approximately 13M tokens, with TRAIL-GAIA traces averaging about 78k tokens each. Edge yields from observational and intervention-validation pipelines on both benchmarks are reported in Table~\ref{tab:graph_construction_yields} and Table~\ref{tab:intervention_yields}.

\begin{table*}[!tbp]
\centering
\footnotesize
\setlength{\tabcolsep}{5pt}
\renewcommand{\arraystretch}{1.1}
\begin{tabular}{l rr}
\toprule
\textbf{Property} & \textbf{TRAIL} & \textbf{MAST} \\
\midrule
Annotated traces      & 148   & 393   \\
\quad GAIA split      & 117   & --    \\
\quad SWE-Bench split & 31    & --    \\
\quad AG2 split       & --    & 393   \\
Leaf error categories & 19    & 13    \\
Annotated error instances & 841 & 1{,}560 \\
\quad GAIA split      & 585   & --    \\
\quad SWE-Bench split & 256   & --    \\
Errors per trace (mean) & 5.7 & 4.0   \\
Corpus size ($\approx$ tokens) & 12.7M & 0.36M \\
Top-level groups      & 3     & 3     \\
Language              & English & English \\
Domain                & Tool usage, code repair & Multi-agent math solving \\
Annotation level      & Span-level & Step-level (we annotate) \\
\bottomrule
\end{tabular}
\caption{Overview of the two benchmarks used in our evaluation. Source benchmarks: TRAIL~\cite{deshpande2025trail} and MAST~\cite{cemri2026multi}. Both contain no human-demographic attributes.}
\label{tab:dataset_overview}
\end{table*}

\paragraph{Domain coverage and intended use.}
TRAIL-GAIA contains traces from general-purpose agent tasks (web search, document understanding, multimodal reasoning). TRAIL-SWE-Bench contains long-context traces from real GitHub-issue code repair (median trace length around 213K characters, with tail traces exceeding 1M). MAST-AG2 contains shorter multi-agent conversational traces. Our use of both benchmarks falls within their original release terms, i.e., research-only evaluation of automatic failure-attribution methods.

\paragraph{Hardware and per-pass compute.}
Open-weight inference runs were executed on a SLURM-managed cluster partition equipped with NVIDIA A100 GPUs. Each evaluation job requested 4 A100 GPUs on a single node with 256 GB of host memory. Closed-source backbones (Gemini-2.5-Pro, Gemini-2.5-Flash, GPT-4o, GPT-5) were accessed through their hosted APIs and do not consume local GPU time. Table~\ref{tab:compute_wallclock} reports the mean per-pass wall-clock and GPU-hour cost. Graph construction and intervention validation are one-time preprocessing steps and are not included in this budget.

\begin{table}[!tbp]
\centering
\footnotesize
\setlength{\tabcolsep}{5pt}
\renewcommand{\arraystretch}{1.1}
\begin{tabular}{l l rr}
\toprule
\textbf{Benchmark} & \textbf{Setting} & \textbf{Wall (min)} & \textbf{GPU Hrs} \\
\midrule
MAST  & Baseline       & 19.2 & 1.28 \\
MAST  & +CG            & 20.2 & 1.35 \\
MAST  & \our (+GI)     & 24.9 & 1.66 \\
\midrule
TRAIL & Baseline       & 48.8 & 3.25 \\
TRAIL & +CG            & 51.2 & 3.41 \\
TRAIL & \our (+GI)     & 78.5 & 5.23 \\
\bottomrule
\end{tabular}
\caption{Mean wall-clock and GPU Hrs per single inference pass on open-weight backbones (GPU Hrs $=$ wall-clock hours $\times 4$, the A100 count per job). MAST means are over Gemma-3-27B-IT, GPT-oss-20B, and GPT-oss-120B (393 traces). TRAIL means are over Gemma-3-27B-IT, GPT-oss-20B, GPT-oss-120B, and Qwen Family on the full corpus (148 traces). +CG matches Baseline in LLM-call count and adds only a small static graph block, so its wall-clock is Baseline plus a $\sim$5\% context-overhead margin. \our (+GI) further adds a trace-conditioned Stage-2 on top of Stage-1, so its wall-clock is strictly higher than +CG.}
\label{tab:compute_wallclock}
\end{table}

\paragraph{LLM serving and inference.}
Open-weight backbones (Mistral-Small-3.1-24B~\cite{mistral2025small31}, GPT-oss-20B and GPT-oss-120B~\cite{openai2025gptoss}, Gemma-3-27B-IT~\cite{gemmateam2025gemma3}, QwenLong-L1-32B~\cite{wan2025qwenlong}, QwQ-32B~\cite{qwen2025qwq}, and Qwen3.6-35B-A3B~\cite{qwen2026qwen36}) are served with \texttt{vllm}\footnote{\url{https://github.com/vllm-project/vllm}} version 0.18.1 using 4-way tensor parallelism, default sampling temperature, and per-model maximum context lengths reported by each model card. Closed-source backbones (GPT-4o~\cite{openai2024gpt4o}, GPT-5~\cite{openai2025gpt5}, Gemini-2.5-Pro and Gemini-2.5-Flash~\cite{comanici2025gemini25}) are accessed through their hosted APIs via \texttt{litellm}\footnote{\url{https://github.com/BerriAI/litellm}}.

\paragraph{Graph construction packages.}
The Suppes screen and CAPRI-style pruning of Section~\ref{candidate_graph} are implemented from scratch in Python rather than via the original R package \texttt{TRONCO}~\cite{desano2016tronco}. The implementation uses \texttt{networkx} (3.4.2) for directed-acyclic-graph manipulation, with custom hill-climbing over edge add/remove/reverse moves and an AIC-style score (see ``Causal Graph Construction Details'' in Appendix~\ref{app:graph_construction}).

\paragraph{Evaluation packages.}
All trace-level and span-level metrics are computed with \texttt{scikit-learn} (1.7.2). Weighted F1, macro precision, macro recall, and macro accuracy are obtained from \texttt{sklearn.metrics} with \texttt{average="weighted"} for the headline F1 and \texttt{average="macro"} for the macro variants.

\end{document}